\documentclass{article} 
\usepackage{iclr2026_conference,times,array, booktabs}

\usepackage{amsmath,amsfonts,bm}

\def\eqref#1{equation~\ref{#1}}

\def\1{\bm{1}}

\DeclareMathAlphabet{\mathsfit}{\encodingdefault}{\sfdefault}{m}{sl}
\SetMathAlphabet{\mathsfit}{bold}{\encodingdefault}{\sfdefault}{bx}{n}

\usepackage{hyperref}
\usepackage{url}
\usepackage{algorithm}
\usepackage{algpseudocode}
\makeatletter
\def\BState{\State\hskip-\ALG@thistlm}
\makeatother
\usepackage{dsfont}
\usepackage{listings}
\usepackage{makecell}
\usepackage[dvipsnames]{xcolor} 
\usepackage{ragged2e}
\usepackage{multirow}
\usepackage{graphicx}
\usepackage{siunitx}
\usepackage{tabularx}
\usepackage[inline]{enumitem}

\usepackage{tcolorbox}

\newenvironment{promptbox}[1]{%
    \begin{tcolorbox}[title={#1}]
}{%
    \end{tcolorbox}
}

\usepackage{booktabs,multirow,siunitx,xcolor,xfp}
\usepackage{siunitx,xcolor}
\usepackage{caption}
\usepackage{amsthm}            
\newtheorem*{theorem*}{Theorem}  

\theoremstyle{plain}
\newtheorem{theorem}{Theorem}[section]      
\newtheorem{lemma}[theorem]{Lemma}          
\newtheorem{proposition}[theorem]{Proposition}

\theoremstyle{definition}

\theoremstyle{remark}

\lstdefinestyle{prompt}{
  basicstyle=\ttfamily\small,
  breaklines=true,
  columns=fullflexible,
  frame=single,
  framerule=0.4pt,
  framesep=6pt,
  rulecolor=\color{black!15},
  backgroundcolor=\color{black!02},
  numbers=none,
  numberstyle=\tiny\color{black!45},
  xleftmargin=1.2em,
  xrightmargin=1.2em,
  aboveskip=0.8\baselineskip,
  belowskip=0.8\baselineskip,
  captionpos=b
}

\title{Unbiased Visual Reasoning with Controlled Visual Inputs}

\author{
\centerline{
Zhaonan Li\textsuperscript{\rm 1*}\qquad
Shijie Lu\textsuperscript{\rm 1*}\qquad
Fei Wang\textsuperscript{\rm 2}\qquad
Jacob Dineen\textsuperscript{\rm 1}\qquad
Xiao Ye\textsuperscript{\rm 1}\qquad
Zhikun Xu\textsuperscript{\rm 1}\qquad
}\\
\centerline{
\textbf{
Siyi Liu\textsuperscript{\rm 3} \qquad
Young Min Cho\textsuperscript{\rm 3} \qquad
Bangzheng Li\textsuperscript{\rm 4} \qquad
Daniel Chang\textsuperscript{\rm 1} \qquad
Kenny Nguyen\textsuperscript{\rm 1} \qquad
}}
\\
\centerline{
\textbf{
Qizheng Yang\textsuperscript{\rm 1}\qquad
Muhao Chen\textsuperscript{\rm 4}\qquad
Ben Zhou\textsuperscript{\rm 1}
}}\\
\centerline{
\textsuperscript{\rm 1}Arizona State University \quad
\textsuperscript{\rm 2}University of Southern California
}\\
\centerline{
\textsuperscript{\rm 3}University of Pennsylvania \quad
\textsuperscript{\rm 4}University of California, Davis\quad
}}

\newif\ifshowrevisions
\showrevisionstrue        

\ifshowrevisions
  
\else
  
\fi

\newcommand{\method}[0]{\textsc{VISTA}} 

\iclrfinalcopy 
\tcbuselibrary{most}
\tcbuselibrary{breakable}

\newcommand{\edit}[1]{#1}
\begin{document}

\maketitle
\thispagestyle{plain}
\pagestyle{plain}  

\begin{abstract}

\edit{
End-to-end Vision-language Models (VLMs) often answer visual questions by exploiting spurious correlations instead of causal visual evidence, and can become more shortcut-prone when fine-tuned. We introduce \textbf{\method{}} (Visual-Information Separation for Text-based Analysis), a modular framework that decouples perception from reasoning via an explicit information bottleneck. A frozen VLM sensor is restricted to short, objective perception queries, while a text-only LLM reasoner decomposes each question, plans queries, and aggregates visual facts in natural language. This controlled interface defines a reward-aligned environment for training unbiased visual reasoning with reinforcement learning. Instantiated with Qwen2.5-VL and Llama3.2-Vision sensors, and trained with GRPO from only 641 curated multi-step questions, \method{} significantly improves robustness to real-world spurious correlations on SpuriVerse (+16.29\% with Qwen-2.5-VL-7B and +6.77\% with Llama-3.2-Vision-11B), while remaining competitive on MMVP and a balanced SeedBench subset. \method{} transfers robustly across unseen VLM sensors and is able to recognize and recover from VLM perception failures. Human analysis further shows that \method{}’s reasoning traces are more neutral, less reliant on spurious attributes, and more explicitly grounded in visual evidence than end-to-end VLM baselines.
}
\end{abstract}

\section{Introduction}
\begin{figure*}[h]
  \centering
  \includegraphics[width=\textwidth, clip, trim={0.75cm 2cm 1.5cm 0.5cm}]{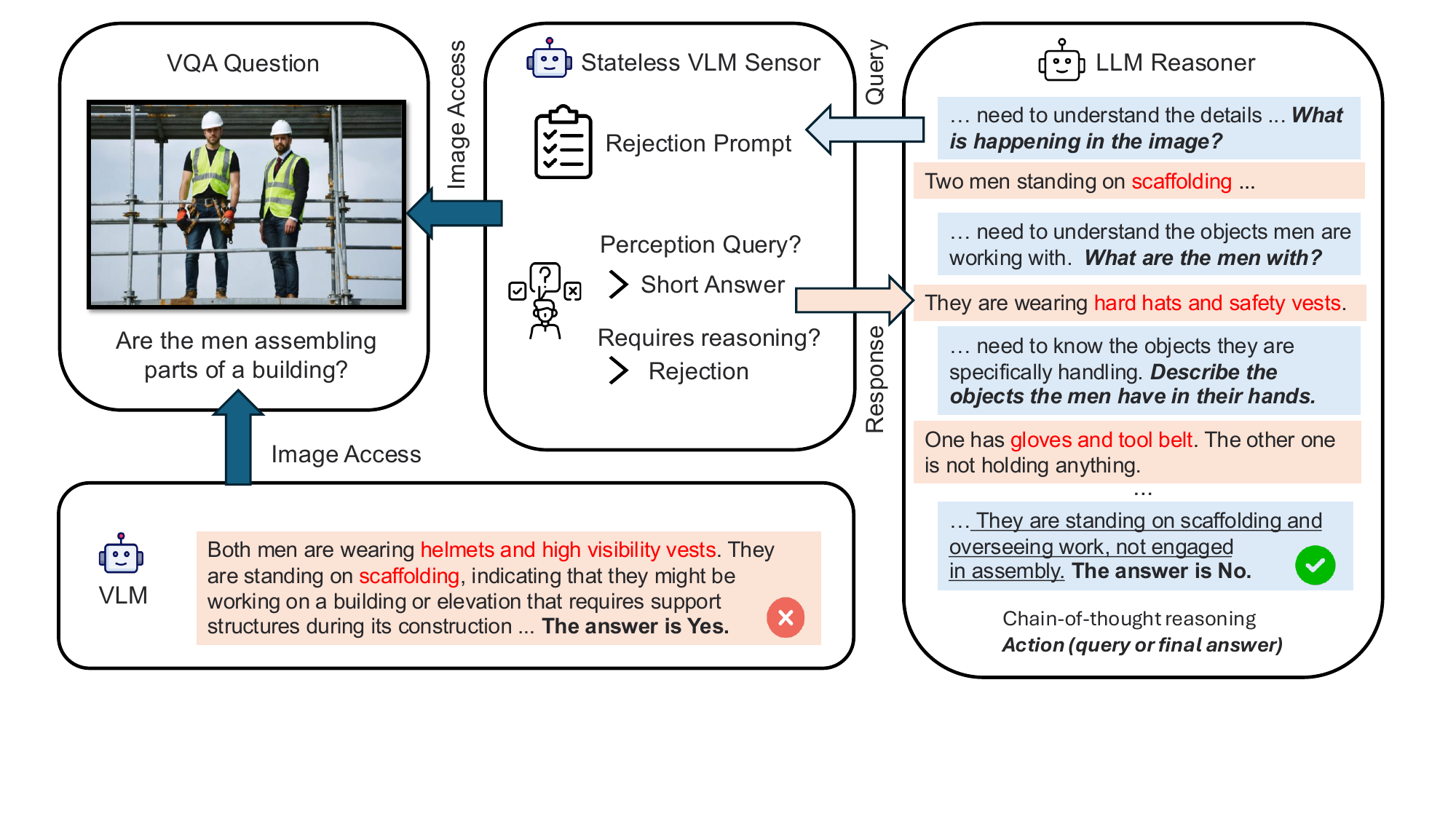}
  \vspace{-1cm}
  \caption{Comparison between an end-to-end VLM and \method{} on a SpuriVerse example (actual model outputs). Spurious attributes are highlighted in red. \textbf{Bottom}: The end-to-end Qwen2.5-VL model predicts \textbf{\emph{Yes}} by exploiting spurious attributes (e.g., scaffolding and stereotypical attire) that are irrelevant to the question, resulting in an error. \textbf{Top}: \method{} decouples perception from reasoning via an information bottleneck and follows a neutral, iterative decision process: the LLM reasoner emits CoT rationales before each action, issues targeted simple visual queries as actions, and terminates the interaction once a conclusion is reached. By explicitly checking the men's actions and interactions, the reasoner remains invariant to the spurious cues and correctly predicts \textbf{\emph{No}}.}
  \label{fig:teaser}
\end{figure*}

Recent advances in vision-language models (VLMs) have propelled multimodal understanding and visual question answering (VQA) to new heights. However, beneath these impressive benchmarks lies a persistent concern: many systems appear to succeed not by genuine visual reasoning, but by exploiting shortcuts that correlate spuriously with the correct answer, including contextual cues, visual predominance, or commonly co-occurring objects \citep{yang2025escaping, kervadec2021roses, dancette2021beyond, si-etal-2022-language, agrawal2018don, wang2024muirbench, wang2024mdpo, ye2024mm}. An example is illustrated in Figure \ref{fig:teaser}: when asked ``are the men assembling parts of a building?'', the end-to-end Qwen2.5-VL-7B model answers ``yes'' based on the presence of scaffolding and stereotypical attire, while failing to verify whether any assembly action is actually taking place. \edit{Beyond benchmarks, similar shortcut behaviors can lead to non-robust and harmful decisions in safety-critical domains such as medical imaging, where models exploit confounding artifacts \citep{degrave2021ai, zech2018variable}, underscoring the importance of evidence-grounded visual reasoning.}

\edit{Critically, this conflation of perception and reasoning is problematic not only at inference but also during training. When a model is trained end-to-end from answers, it is difficult to provide learning signals that distinguish relevant causal evidence from correlated but irrelevant cues. This pattern is consistent with a broader literature on shortcut learning, which shows that deep networks preferentially latch onto easily available but non-causal cues \citep{geirhos2020shortcut}. Recent advances in VLM reasoning via Supervised Fine-tuning (SFT) or Reinforcement Learning (RL) with verifiable reward do not fundamentally alter this picture: RL-trained models can be susceptible to reward hacking and lack fine-grained supervision \citep{gao2023scaling, lightman2023let}, and SFT can capture spurious associations between particular visual patterns and textual attributes \citep{varma2024ravl}.
These findings suggest that simply applying SFT or RL to end-to-end VLMs cannot be expected to automatically discover neutral, evidence-seeking visual reasoning and can, in fact, reinforce stereotyped shortcuts and reduce robustness under distribution shift or adversarial conditions.
}

To address these limitations, we introduce \method{} (\textbf{V}isual-\textbf{I}nformation \textbf{S}eparation for \textbf{T}ext-based \textbf{A}nalysis), a new meta-architecture and learning paradigm that enforces an information bottleneck between perception and reasoning modules \citep{alayrac2022flamingo, li2023blip}. Our key intuition is simple: by restricting the reasoning module to obtain visual information only through a limited-information, verifiable query interface, we reduce the opportunity to exploit spurious correlations, force the reasoner to rely on logical deduction as well as targeted evidence gathering, and create a clearer credit assignment during training. In \method{}, raw visual inputs are handled by a frozen VLM that answers simple and objective queries, while rejecting complex or vague requests. All decision-making occurs in an LLM reasoner that plans queries, integrates visual evidence, and explains its conclusions step-by-step. As illustrated in Figure~\ref{fig:teaser}, \method{} operates as an iterative dialogue between an LLM reasoner and a VLM sensor constrained to answer simple, objective queries. Given a question, the reasoner conducts step-wise, verbalized planning and issues targeted visual checks, while the sensor either rejects or returns short factual responses. In the earlier example, our reasoner remains unaffected by spurious attributes and explicitly verifies the men's actions. By pursuing a neutral, evidence-seeking reasoning path, \method{} concludes with the correct answer. 

\edit{To investigate the effectiveness of our information-bottleneck design, we deliberately restrict our scope to common-sense VQA on natural images of everyday scenes. In this regime, visual evidence, such as objects, attributes, spatial relationships, and activities, can be faithfully translated into short textual facts via simple perception queries, making the perception–reasoning separation identifiable and minimizing information loss through the bottleneck. By contrast, chart- and table-based VQA, as well as domain-specific math or science VQA, often require (1) holistic, non-local interpretations (e.g., reading a derivative from a curve or aggregating across axes) that are difficult to decompose into atomic perception primitives, or (2) external domain knowledge that entangles knowledge gaps with reasoning errors. Evaluating in these settings would confound our core question with orthogonal challenges in specialized parsing and knowledge retrieval. Everyday scenes are also rich in spurious cues (co-occurring objects, stereotyped contexts), making them a natural testbed for reasoning neutrality. Within this scope, our approach offers a proof-of-concept foundation for future adaptation to more complex domains.}

We summarize our contributions as follows:
\begin{itemize}
    \item We propose \method{}, a framework and corresponding learning paradigm that formalizes VQA as an iterative decision-making process under an information bottleneck that separates perception from reasoning. 
    \item We demonstrate that, with the same data and training steps, \method{} encourages neutral, evidence-seeking reasoning across two vision backbones, whereas end-to-end training, including SFT and RL, on VLMs reinforces visual shortcuts and reduces robustness.
    \item \method{} attains substantial robustness gains on SpuriVerse while remaining on par with end-to-end systems on everyday-scene benchmarks (MMVP, SeedBench).
\end{itemize}

\section{Related Work}
\noindent \textbf{Modular VQA Systems}. Early modular VQA systems explicitly decompose problems into perception and reasoning components. Neural Module Networks dynamically compose modular networks depending on the question structure \citep{Andreas_2016_CVPR}. Neural-Symbolic VQA parses questions into executable programs against structured scene graphs \citep{yi2018neural}. These methods separate recognition from symbolic reasoning but often rely on strong supervision or curated representations. Later ViperGPT and VisProg show that LLMs, with strong built-in code generation capabilities, can compose visual operators as programs, offering strong interpretability and compositional generalization \citep{suris2023vipergpt, gupta2023visual}. Compared with these programmatic modular systems, our formulation uses language as the interface to perception, avoiding coverage gaps and engineering constraints imposed by APIs or program libraries. In addition, our reasoning proceeds iteratively, which supports complex reasoning and produces auditable traces. Crucially, we impose an information bottleneck to mitigate visual biases, which underpins our motivation to encourage neutral visual reasoning. To address limitations from domain-specific decomposition and premature conclusions without sufficient visual information in multi-step VQA, IdealGPT decomposes questions into sub-questions and delegates answering to a VLM \citep{you2023idealgpt}. Our formulation shares the same high-level recipe, including LLM-based decomposition and iterative reasoning, but differs in fundamental ways: (1) we enforce a perception-only interface that explicitly targets visual bias mitigation; (2) we study a training paradigm and compare directly with end-to-end VLM training, whereas IdealGPT is evaluated zero-shot with a closed LLM; (3) our method trains a single LLM to decompose, reason, and decide the final answer, whereas IdealGPT assumes separate strong models (ChatGPT) for questioning and reasoning.


\noindent \textbf{Robustness, Shortcut Learning, and Evaluation Benchmarks}. VQA robustness work shows that models often exploit shortcuts rather than genuine reasoning. VQA-CP introduces changing-prior splits to break question-type priors and reveals large drops for models under shifted priors \citep{agrawal2018don}. Beyond question-only biases, VQA-CE mines multimodal shortcut rules and demonstrates that many debiasing methods remain ineffective when the shortcuts are cross-modal \citep{dancette2021beyond}. GQA-OOD reorganizes the GQA dataset and finds that strong VQA models still fail on infrequent or shifted compositions \citep{kervadec2021roses}. More recently, MM-SpuBench probes spurious biases by asking models to pick the diagnostic feature for object identity \citep{ye2024mm}. Since our claims center on QA accuracy under controlled spurious shifts and reasoning, we consider datasets aligned with those goals. SpuriVerse curates real-world VLM failures attributed to spurious cues and validates them with synthetic counterfactuals \citep{yang2025escaping}. In parallel, MMVP targets basic visual-pattern failures and SeedBench provides broad, human-annotated multiple-choice evaluations and enables standardized comparison across models \citep{tong2024eyes, li2023seed, li2024seed}. Our approach is complementary to dataset-level and loss-level debiasing: instead of reweighting data or adding regularizers, we enforce an architectural bottleneck that promotes neutral visual reasoning while remaining compatible with everyday suites and spurious-stress evaluations.

\noindent \textbf{Active Reasoning and Reinforcement Learning}. 
Active information-seeking has been studied in multi-hop QA and fact verification \citep{yang2018hotpotqa, thorne2018fever} as well as in interactive environments \citep{shridhar2020alfworld, yao2022webshop, zhou2023webarena}. LLM agents often alternate between planning, tool use, and verification, sometimes under explicit budgets. Foundational systems interleave reasoning with actions \citep{yao2023react}, browse and cite sources with human feedback \citep{nakano2021webgpt}, and improve over trials via self-reflection \citep{shinn2023reflexion}. Our setting shares the multi-turn nature but differs in objective: rather than maximizing task success by any means, we explicitly constrain how information can be acquired to prevent shortcut learning.

On learning signals, RL has been effective for aligning multi-turn behaviors and tool use. Popular training paradigms include PPO-based RLHF with KL control for long-horizon tool use and dialogue \citep{nakano2021webgpt, ouyang2022training}, AI-feedback variants that reduce human labeling \citep{bai2022constitutional, lee2023rlaif}, and offline preference optimization \citep{rafailov2023direct}. Recent group-based objectives (GRPO) stabilize reasoning-centric training by scoring multiple completions per prompt and using relative advantages \citep{shao2024deepseekmath}. Our setting is algorithm-agnostic, and we adopt GRPO for its practicality and strong uptake in reasoning-focused LLMs.

\begin{algorithm}[t]
\caption{\method{} reasoning loop}
\label{alg:blind}
\begin{algorithmic}[1]
\Require image $x$, question $q$, reasoner $\pi_\theta$, sensor $S_\phi$, step budget $T_{\max}$
\State $h \gets [\,q\,]$ \Comment{Reasoner history of pairs $(s_i, y_i)$}
\For{$t = 1$ \textbf{to} $T_{\max}$}
    \State $s_t \sim \pi_\theta(\cdot \mid h)$ \Comment{Reasoner raw text at step $t$}
    \State parse $s_t \rightarrow (c_t, u_t)$
    \If{$u_t = \mathrm{ANSWER}(a_t)$}
        \State \Return $a_t$ \Comment{Terminate upon answer}
    \ElsIf{$u_t = \mathrm{QUERY}(q_t)$}
        \State $y_t \gets S_\phi(x, q_t)$
        \Comment{Sensor sees only $(x, q_t)$; no $q$, options, or history}
        \State $h \gets h \,\|\, (s_t, y_t)$ \Comment{Append $(s_t, y_t)$ to history}
    \EndIf
\EndFor
\State \Return $s_t$
\end{algorithmic}
\end{algorithm}

\section{Method}
\subsection{Overview}
 We decompose a VQA system into a text-only \textbf{reasoner} $\pi_{\theta}$, and a frozen VLM \textbf{sensor} $S_{\phi}$ that answers perception-only questions. Given the textual input question $q$, the reasoner iteratively interacts with the sensor by issuing free-form natural-language queries; the sensor sees the input image $x$, and either returns a short answer or rejects the query when it requires high-level inference. The interaction loop terminates when the reasoner concludes with an answer or the maximum number of steps is reached.

Formally, at step $t$, the reasoner observes the conversation history
$$
h_t = \bigl( q, (s_1, y_1), \ldots , (s_{t-1}, y_{t-1}) \bigr),
$$
where $s_i$ and $y_i$ are the output strings of $\pi_{\theta}$ and $S_{\phi}$ at step $i$. Each $s_t$ contains two parts:
\begin{itemize}
    \item Chain-of-thought $c_t$: text used by the reasoner to think before outputting an action
    \item Action $u_t$: a structured directive extracted with a deterministic rule-based parser
\end{itemize}
The reasoner implicitly learns to decide \emph{what to ask} and \emph{when to stop}. The action space is
$$
    u_t \in \mathcal{U} = \{ \texttt{QUERY}(q_t), \texttt{ANSWER}(a_t)\},
$$
where $q_t$ is a query string and $a_t$ is a final answer string. The loop terminates if a final answer is extracted; otherwise, the sensor receives $q_t$ and returns
$$
    y_t = S_{\phi}(x, q_t).
$$ Crucially, the stateless sensor $S_{\phi}$ never receives the history or the reasoning traces. It sees only the contextless query $q_t$ and image $x$. Thus, all task-level decision-making must arise from $\pi_{\theta}$. The working pipeline of \method{} is illustrated in Algorithm \ref{alg:blind}.


\subsection{Perception-Only Sensor and Rejection Policy}\label{sec:perception-queries}

\paragraph{Perception-only sensor.}
\edit{Our design goal is to delegate as much high-level reasoning as possible to the text-only LLM reasoner and constrain the VLM to act as a simple, factual sensor. To obtain a clean separation between perception and reasoning, we restrict the sensor to answer only perception queries that fall into the taxonomy proposed by \citet{Selvaraju_2020_CVPR}: \emph{Existence} (``Is there a bicycle?''), \emph{Basic Properties} (``Is the mug red?''), \emph{Spatial Relations} (``What is left of the sofa?''), \emph{Simple Activities} (``Are they looking at the camera?''), \emph{Text/Symbol Recognition} (``What does the road sign say?''), and \emph{Counting} (``How many cups are on the table?''). Within \method{}, the VLM sensor is therefore responsible only for reporting such low-level visual facts, while all decomposition, integration of evidence, and high-level decision-making are handled by the LLM reasoner. To help the reasoner when the question lacks sufficient context, the sensor may optionally return a single, short, objective \emph{overview} describing only scene type, dominant objects, layout, and basic global attributes.}

\begin{figure*}[t]
  \centering
  \includegraphics[width=\textwidth, clip, trim={3cm 4cm 2cm 2.9cm}]{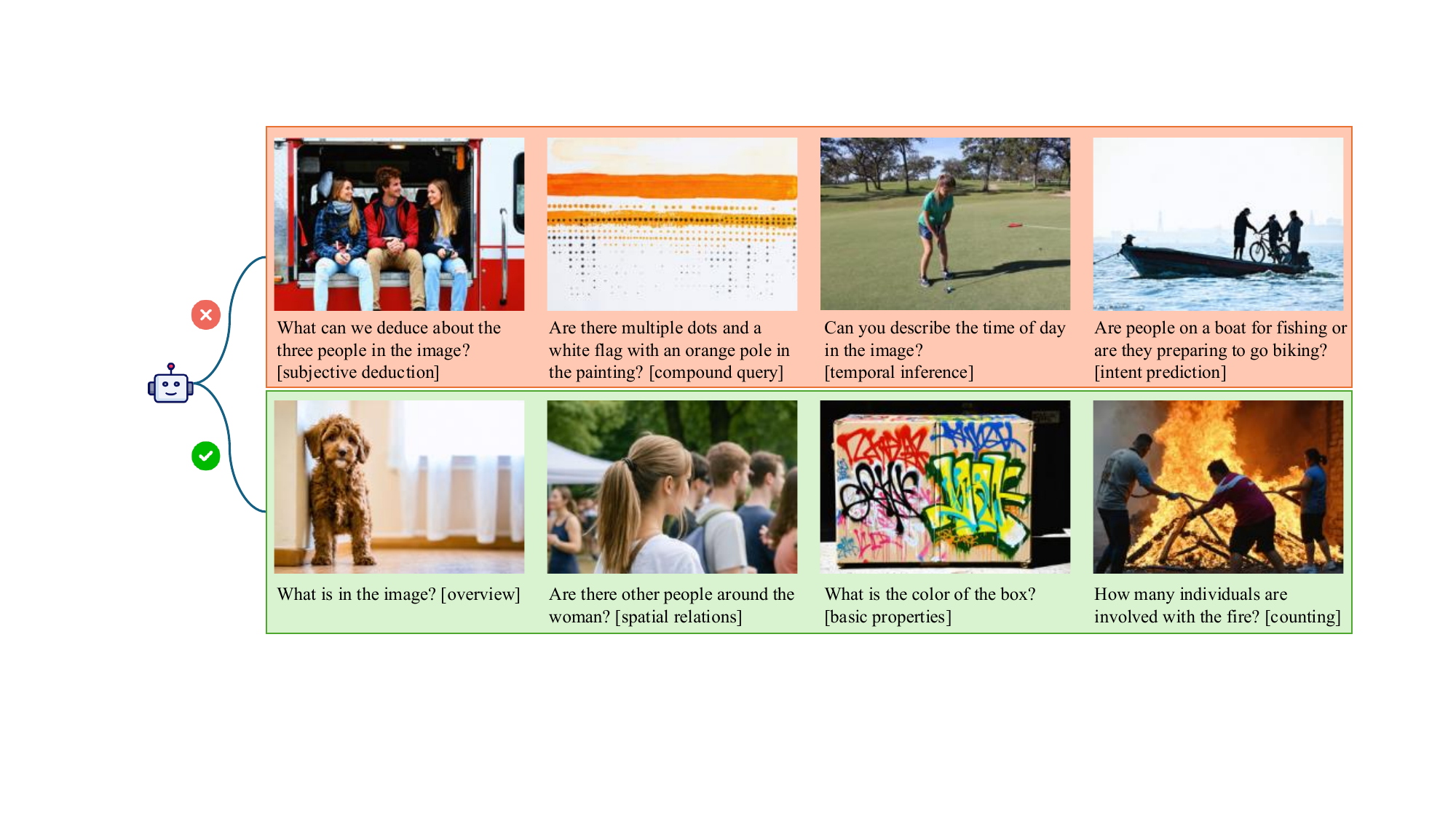}
  \vspace{-0.8cm}
  \caption{\textbf{Accepted vs.\ rejected queries.} The top row shows rejected cases, and the bottom row shows accepted cases. The vision-only \emph{sensor} answers perception questions in six categories and may emit one brief \textsc{Overview} when the text is under-specified; all requests requiring high-level inference are \textsc{Rejected}.}
  \label{fig:accep_reject_queries}
\end{figure*}

\paragraph{Rejection policy and enforcement.}
\edit{Any request that goes beyond these perception categories, requires multi-hop or causal reasoning, external knowledge, or subjective interpretation, bundles multiple inferences instead of simple primitives, or remains ambiguous even after an overview must be \emph{rejected} with the fixed response: ``I cannot answer this question.'' We implement this behavior via an explicit accept/reject instruction prompt and a constrained response format, so that the reasoner is forced to rephrase high-level or underspecified requests into simpler perception queries. Examples of accepted and rejected queries are shown in Fig.~\ref{fig:accep_reject_queries}, and the full prompt is given in Appx.~\ref{app:prompt}. Human analysis (Section~\ref{man_analysis}) of 100 random cases shows 86\% agreement with human pass/reject labels, indicating that this rejection policy is reliably followed.
}




\subsection{Reinforcement Learning Reasoner} Our learning strategy formulates VQA solving as a sequential decision-making process and provides the reasoner with an explorable environment with clear reward signals, making RL training a well-suited choice. We optimize $\pi_{\theta}$ using Group Relative Policy Optimization (GRPO)~\citep{shao2024deepseekmath}.
Each episode $\tau$ yields a terminal reward based on final answer correctness:
\begin{equation}
    R(\tau)=\mathds{1}[a_T=a^*].
\end{equation}

Training differs from single-step GRPO only in the sampling of rollouts and the assignment of loss masks.
We apply the GRPO update to the union of \emph{assistant-only} tokens across all assistant turns.
With terminal-only reward and unit discount, the group-relative advantage is constant within a trajectory, so the update is effectively the single-step GRPO objective applied to a longer, state-dependent sequence (details in App.~\ref{app:multiturn-grpo}).

\section{Theoretical Analysis}
Intuitively, overfitting thrives when the learner can absorb rich, high-variance signals and latch onto spurious correlations that happen to predict labels in the training set. By constraining the visual bandwidth, we shrink the hypothesis space the reasoner can realize: high-level, shortcut features cannot pass through the interface, forcing predictions to rest on a small set of stable, perception-level facts. In this section, we formalize this intuition by relating generalization to the information that can flow through the sensor–reasoner interface.
\paragraph{Setup.}
Let $\bigl(X, Q, Y\bigr) \sim D$ denote image, question, label. A reasoner interacts with a sensor for at most $T$ steps. At step $t$, the reasoner emits a free-form text query $a_t$; the sensor enforces a rejection rule $R_t = g(a_t) \in \{0,1\}$: if $R_t = 0$, it returns a rejection template $\perp$; otherwise returns a short perception answer from a finite alphabet $O_t \in \Sigma$. Let $Z_{1:T} = \bigl(Z_1, \ldots, Z_{T} \bigr), Z_{t} \in \Sigma_{\perp} := \Sigma \cup \{\perp\} $ be the visual evidence. We train parameters $W$ from the compressed dataset $\tilde{D} = \{(Z_{1:T}, Q_i, Y_i)\}_{i=1}^{n}$. We assume the learning loss $\ell(W; Z, Q, Y) \in [0,1]$ is bounded. The true loss and empirical loss are defined as $L(W) = \mathbb{E}\ell(W; Z, Q, Y)$ and $\hat{L}(W, \tilde{D}) = \frac{1}{n} \sum_{i=1}^{n}\ell(W; Z_i, Q_i, Y_i)$.

\begin{theorem*}[Informal, generalization under an information bottleneck]
$$
|\mathbb{E}[\hat{L}(W, \tilde{D}) - L(W)]| \leq \sqrt{2C_T},
$$
where $C_T$ is the per-example bit budget
$$
    C_T := T \log |\Sigma_{\perp}|
$$
\end{theorem*}

\paragraph{Implications and Limitations.}
The expected generalization gap depends only on the interface budget $C_T$ and is independent of the size of the training data, where a smaller $C_T$ means less overfitting. While the bound captures average generalization, it does not alone guarantee worst-case adversarial robustness nor account for distribution shift without extra assumptions. The complete proof is included in Appendix \ref{sec:theory}.

\section{Experiment Setup}
\subsection{Datasets and Preprocessing}
 \edit{We evaluate on three multiple-choice VQA benchmarks over natural images, chosen to match our scope of common-sense VQA on everyday scenes and with no question overlap with our training set}: SpuriVerse \citep{yang2025escaping}, MMVP \citep{tong2024eyes}, and SeedBench \citep{li2023seed}. SpuriVerse consists of 1200 questions explicitly constructed around real-world spurious correlations, making it well-suited for testing reasoning robustness under adversarial conditions. MMVP stresses perceptual limitations by constructing CLIP-blind image pairs and associated questions that expose visual-grounding failures. SeedBench is for everyday, non-adversarial performance. \edit{To obtain a balanced, representative subset, we sample up to 100 questions from each of the nine image-related categories (including all questions for categories with fewer than 100), yielding 882 instances in total}. Because in SpuriVerse more than $60\%$ of gold answers appear in option B, we mitigate answer-position bias by shuffling the multiple-choice options. Shuffling is applied once as a deterministic pre-processing step, and the exact same shuffled inputs are used across all evaluation settings. We report both the original and shuffled results in Appendix \ref{app:shuffle} and observe that our method consistently outperforms all baselines and yields significant improvements. We present the shuffled results in the main text, as they remove label-position bias while preserving the overall trend.

\subsection{\method{} and baseline settings}
\noindent\textbf{\method{}}. For all experiments, we use Qwen2.5-7B as the LLM reasoner. We train and instantiate our method with two frozen VLM sensors: Qwen2.5-VL-7B and Llama3.2-11B. 
For each sensor, we evaluate two settings: 
\begin{enumerate*}[label=(\roman*), itemjoin={{; }}, itemjoin*={{; and }}]
\item \method{} (base): with an untrained reasoner (reference model) interacting with the sensor
\item \method{} (RL): with trained reasoner using GRPO
\end{enumerate*}.

\noindent\textbf{Baselines}. We compare against end-to-end VLMs using the same two backbones in the following settings:
\begin{enumerate*}[label=(\roman*), itemjoin={{; }}, itemjoin*={{; and }}]
    \item E2E (base): the untrained VLM directly answers the question
    \item E2E (base + CoT): the untrained VLM outputs chain-of-thoughts before answers
    \item E2E (SFT): supervised fine-tuning to directly answer
    \item E2E (RL): we additionally evaluate a GRPO-trained Qwen2.5-VL-7B on the same training data and for the same number of steps as \method{} (RL)
\end{enumerate*}.
These baselines isolate where gains come from in our framework design and training signals.

\subsection{Evaluation Protocols}
We report accuracy on SpuriVerse, MMVP, and \edit{SeedBench-Avg}. For a fair comparison, we standardize sampling and decoding across methods: both \method{} and end-to-end VLMs use 11-sample self-consistency at temperature~1.0 for the \emph{predictive component} (the LLM reasoner in \method{} and the VLM itself in end-to-end baselines), and the majority-voted answers are evaluated. For \method{}, the reasoner–sensor interaction is capped at $T_{\max}=24$ and the LLM reasoner is sampled at temperature~1.0, while the VLM sensor's temperature is set to 0 during both training and evaluation. Because end-to-end VLMs may emit unparsable multiple-choice strings, we canonicalize raw outputs to the option set with a lightweight Qwen-2.5-7B post-processor prior to evaluation.

\subsection{Training Setup}
We construct the training set by sampling questions from five sources: VQAv2~\citep{goyal2017making}, Visual7W~\citep{zhu2016visual7w}, GQA~\citep{ainslie2023gqa}, A-OKVQA~\citep{schwenk2022okvqa}, and VQA-Introspect~\citep{Selvaraju_2020_CVPR}. We then apply a multi-stage filtering pipeline that (1) retains questions likely to elicit multi-step reasoning and (2) removes examples solvable via easy visual or textual shortcuts. This yields a training split of 641 questions(A-OKVQA: 502, VQA-Introspect: 95, Visual7W: 34, VQAv2: 7, GQA: 3). Details of the filtering process and the resulting composition are summarized in Appendix \ref{app:data_comp}. We provide details of RL and SFT training in the Appendix \ref{app:training_details}.

\section{Main Results}

\newcommand{\gnum}[1]{\textcolor{green!50!black}{#1}}
\newcommand{\rnum}[1]{\textcolor{red!60!black}{#1}}
\newcommand{\best}[1]{\bfseries #1}

\begin{table}[t]
\centering
\small
\setlength{\tabcolsep}{6pt}
\renewcommand{\arraystretch}{1.1}
\caption{Main results on SpuriVerse, MMVP and SeedBench-Avg. }
\label{tab:main_results}
\begin{tabular}{ll
                S[table-format=2.2]
                S[table-format=2.2]
                S[table-format=2.2]
                S[table-format=2.2]
                S[table-format=2.2]
                S[table-format=2.2]
                }
\toprule
\textbf{VLM} & \textbf{Setting} & \textbf{SpuriVerse} & $\Delta$ & \textbf{MMVP} & $\Delta$ & \textbf{SeedBench-Avg} & $\Delta$ \\
\midrule
\multirow{7}{*}{\shortstack[c]{Qwen2.5-\\VL}}
  & E2E (base)        & 37.50 & & 51.33 & & 75.31 & \\
  & E2E (base + CoT)  & 47.42 & \gnum{+9.92} & 52.67 & \gnum{+1.34} & 77.60 & \gnum{+2.29} \\
  & E2E (SFT)         & 34.84 & \rnum{-2.66} & 50.67 & \rnum{-0.66} & 75.67 &  \gnum{+0.36} \\
  & E2E (RL)          & 44.52 & \gnum{+7.02} & \best{53.33} & \gnum{+2.00} & \best{77.96} &  \gnum{+2.65} \\
  \cmidrule(lr){2-8}
  & \method{} (base)  & 46.29 & \gnum{+8.79} & 46.67 & \rnum{-4.66} & 68.86 & \rnum{-6.45} \\
  & \method{} (RL)    & \best{53.79} & \gnum{+16.29} & 50.00 &  \rnum{-1.33} & 76.40 & \gnum{+1.09} \\
\midrule
\multirow{6}{*}{\shortstack[c]{Llama3.2-\\Vision}}
  & E2E (base)        & 39.76 & & 45.33 & & 73.34 & \\
  & E2E (base + CoT)  & 38.87 & \rnum{-0.89} & 48.00 & \gnum{+2.67} & 74.77 & \gnum{+1.43} \\
  & E2E (SFT)         & 40.16 & \gnum{+0.40} & 32.00 & \rnum{-13.33} & 68.18 & \rnum{-5.16} \\
  \cmidrule(lr){2-8}
  & \method{} (base)  & 44.44 & \gnum{+4.68} & 35.33 & \rnum{-10.00} & 66.19 & \rnum{-7.15} \\
  & \method{} (RL)    & \best{46.53} & \gnum{+6.77} & \best{52.67} & \gnum{+7.34} & \best{75.21} & \gnum{+1.87} \\
\bottomrule
\end{tabular}
\end{table}

We present our main results in Table \ref{tab:main_results}. We report accuracy on SpuriVerse, MMVP and \edit{SeedBench-Avg}. For each vision backbone, we show the $\Delta$ relative to its corresponding E2E (base); positive changes are highlighted in green and drops in red. The best numbers for each dataset and backbone are bolded.

\noindent\textbf{Robustness to spurious correlations}. We evaluate on SpuriVerse, which is based on real-world spurious cues, and compare our approach with E2E VLM baselines under an identical evaluation protocol. In the inference-only setting (\method{} base in the table), we use an untrained LLM paired with a frozen VLM sensor and our results already match or surpass the best performing E2E systems. With Qwen2.5-VL as the sensor, \method{} scores 46.29\%, approaching the best E2E baseline (untrained + CoT) at 47.42\%; with Llama-3.2-Vision, \method{} reaches 44.44\%, outperforming the best E2E baseline (SFT) at 40.16\%. These results support our design that constraining the interface to perception-only queries keeps the reasoner on a neutral, evidence-seeking path rather than following spurious visual shortcuts, and the gains hold model-agnostically across sensors. With RL-trained reasoners (sensors remain frozen), performance further improves and the gaps widen. On Qwen2.5-VL, RL yields a 7.5\% improvement over our base policy to 53.79\%, extending the margin over the best E2E baseline to 6.37\%; on Llama-3.2-Vision, RL attains 46.53\% and maintains a 6.37\% lead over the strongest E2E (SFT) baseline. We further conduct a manual analysis showing that our gains are associated with more neutral, evidence-linked reasoning traces. Details are in Section \ref{man_analysis}.

\noindent\textbf{General performance on MMVP and SeedBench}. To contextualize robustness results, we evaluate on MMVP and \edit{SeedBench-Avg}, targeting everyday-scene questions whose answers can be inferred from a small set of observable visual predicates combined with commonsense and short multi-step reasoning. Overall, \method{} delivers substantial robustness gains with only marginal accuracy trade-offs relative to the strongest E2E baselines. On MMVP, our RL-trained reasoner improves over the strongest E2E baseline with Llama3.2-Vision (52.67\% vs. 48.00\%) and is only marginal behind the strongest baselines with Qwen2.5-VL by 3.33\%. SeedBench provides a general, non-adversarial testbed. \edit{Our method surpasses the strongest E2E baseline with Llama3.2-Vision (75.21\% vs. 74.77\%) and is slightly below the best E2E baselines with Qwen2.5-VL by 1.56\%. A detailed breakdown by SeedBench-Avg category is provided in Table \ref{tab:seedbench_avg_split} in Appendix \ref{additional-results}.} Because SeedBench does not target adversarial spuriousness, end-to-end VLMs with raw-pixel access can exploit benign correlations and holistic cues, yielding a small but consistent edge. By contrast, our architecture enforces a perception-only interface that promotes neutral, evidence-based reasoning under constrained visual bandwidth, introducing an explicit trade-off between information bandwidth and neutrality. The rejection ablation in Section~\ref{analysis:rejection} supports this hypothesis, and we approach E2E results when the rejection bottleneck is removed.

\begin{figure*}[t]
  \centering
  \includegraphics[width=\textwidth, clip, trim={0.01cm 0.01cm 0.01cm 0.01cm}]{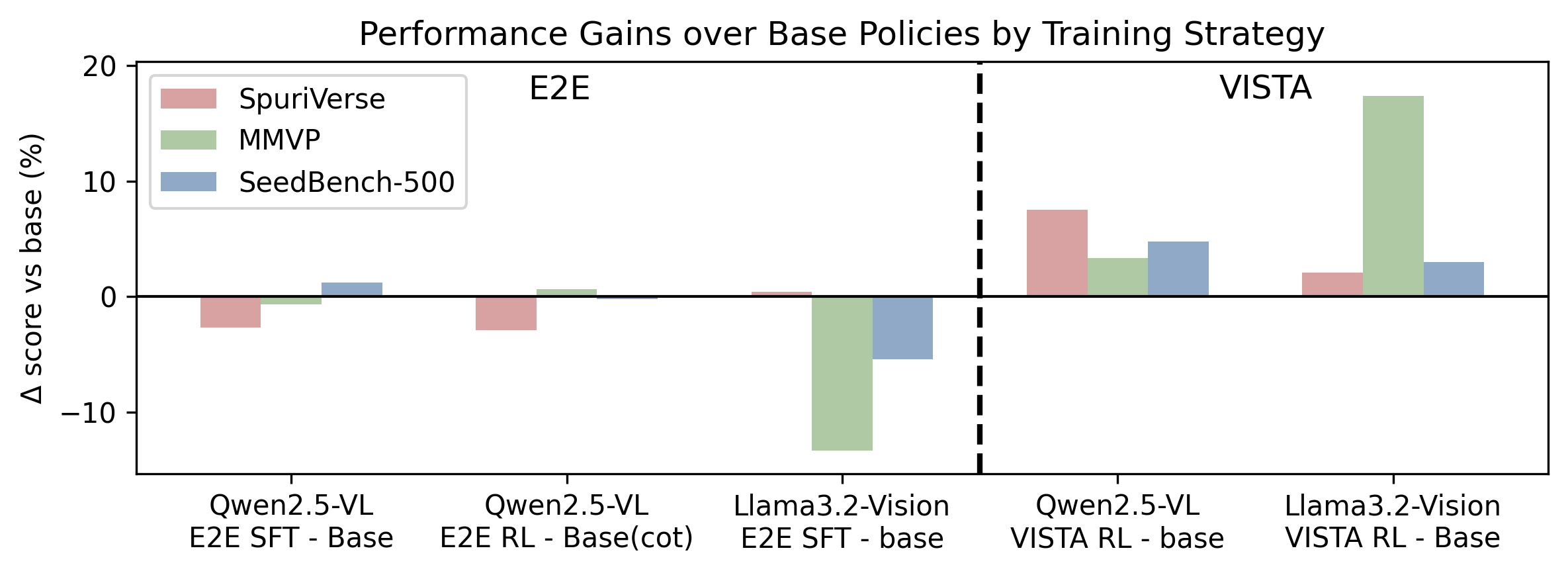}
  \vspace{-0.5cm}
  \caption{Bars show the change in percentage points of each trained model relative to its base policy on three robustness benchmarks. \method{} with RL yields consistent, sizable robustness gains, whereas end-to-end gives negligible or even negative improvements.}
  \label{fig:learning_strategies}
\end{figure*}

\noindent\textbf{Comparison of learning strategies}. We compare SFT and RL applied either to end-to-end VLMs or to our reasoner in \method{}, using the same training data and schedule. In this section, we compare and report the improvement gains of the trained model compared with its base policy, \edit{as shown in Fig.~\ref{fig:learning_strategies}}. For example, E2E SFT baselines are measured against E2E base (no CoT), while E2E RL are measured against E2E base + CoT; \method{} deltas are measured against their own base policy. Across both vision backbones, training \method{} yields consistent, sizable gains over its base, whereas training the VLM end-to-end produces marginal and often inconsistent improvements. The effect is most pronounced on SpuriVerse: Qwen2.5-VL E2E training hurts robustness (SFT -2.66\%; RL -2.90\%); Llama3.2 E2E SFT has a marginal +0.40\% difference). \method{}-RL improves markedly (+7.50\% with Qwen2.5-VL; +2.09\% with Llama3.2-Vision). On MMVP and \edit{SeedBench-Avg}, E2E training yields at best small gains, despite becoming more susceptible to spurious cues as evidenced by the SpuriVerse results. Taken together, these findings indicate that conflating perception and reasoning during E2E training blurs learning signals between causal evidence and correlated but irrelevant features, whereas \method{}’s perception-only interface creates a better-suited learning environment in which RL can reliably shape neutral, evidence-seeking policies.

\section{Analysis and Discussions}

\begin{table}[t]
\centering
\small
\setlength{\tabcolsep}{6pt}
\renewcommand{\arraystretch}{1.1}
\caption{Ablation on the VLM rejection bottleneck with Metrics: acc = accuracy, rnd = average conversation rounds, rej = rejection rate.}
\label{tab:ablation_rejection}
\begin{tabular}{ll
                S[table-format=2.2]  
                S[table-format=1.2]  
                S[table-format=1.2]  
                c                    
                S[table-format=2.2]  
                S[table-format=1.2]  
                S[table-format=1.2]  
                }
\toprule
\textbf{VLM} & \textbf{Setting}
& \multicolumn{3}{c}{\textbf{SpuriVerse}}
& &
\multicolumn{3}{c}{\textbf{SeedBench-Avg}} \\
\cmidrule(lr){3-5} \cmidrule(lr){7-9}
& & \textbf{acc} & \textbf{rnd} & \textbf{rej}
  & & \textbf{acc} & \textbf{rnd} & \textbf{rej} \\
\midrule
\multirow{4}{*}{Qwen2.5-VL}
  & \method{} (base), w/ rejection   & \textbf{46.29} & 3.38 & 0.18 & & 68.86 & 3.43 & 0.20 \\
  & \method{} (base), w/o rejection  & 43.23 & 3.05 & 0.00 & & \textbf{71.02} & 3.03 & 0.00 \\
  \cmidrule(lr){2-9}
  & \method{} (RL), w/ rejection     & \textbf{53.79} & 7.31 & 0.32 & & 76.40 & 6.58 & 0.29 \\
  & \method{} (RL), w/o rejection    & 51.37 & 6.00 & 0.00 & & \textbf{77.33} & 5.42 & 0.00 \\
\bottomrule
\end{tabular}
\end{table}

\subsection{Rejection ablation}\label{analysis:rejection}
We ablate the rejection bottleneck and investigate its effect in two regimes: adversarial spurious correlations (SpuriVerse) and non-adversarial everyday scenes \edit{(SeedBench-Avg)}. The results reveal a clear information-bandwidth–neutrality trade-off. With rejection on, the sensor denies high-level inferences and answers only perception-level queries, shifting the burden to the LLM and encouraging evidence-based reasoning under reduced visual bandwidth. With rejection off, the sensor answers high-level queries, increasing bandwidth but exposing the system to shortcut exploitation. Table~\ref{tab:ablation_rejection} reports accuracy alongside mean conversation rounds and rejection rates for \method{} (base) and \method{} (RL) with/without rejection. Enforcing the bottleneck improves robustness on SpuriVerse, confirming its value for shielding against spurious cues; removing the bottleneck improves \edit{SeedBench-Avg} performance, shortens interactions (fewer rounds), and drives the rejection rate to zero. Notably, the RL variant without rejection attains near-parity with the strongest E2E baseline on \edit{SeedBench-Avg}, suggesting that relaxing the gate can recover benign, non-adversarial cues while the full bottleneck remains preferable under adversarial conditions. Our results also indicate that RL training promotes deeper evidence-seeking, as evidenced by an increase in the average number of conversation rounds. As future work, we will investigate rejection-aware, efficiency-regularized learning to induce more concise reasoning and develop adaptive, confidence-aware gating that modulates rejection to balance information bandwidth and neutrality.

\subsection{Zero-shot Generalization on unseen VLM sensor}
To test whether the policy exploits VLM-specific patterns, we perform a zero-shot sensor swap: the reasoner trained with a Qwen2.5-VL sensor is paired with an unseen Gemma3 sensor. Without any additional tuning, the resulting learned agent substantially improves over the base policy and clearly strengthens robustness on SpuriVerse, while outperforming untrained end-to-end Gemma baselines on MMVP. These trends suggest that the learned reasoning policy largely transfers across vision backbones and that much of the robustness gain comes from sensor-agnostic reasoning. The results are summarized in Table~\ref{tab:zero_shot}.
\begin{table}[t]
\centering
\small
\setlength{\tabcolsep}{6pt}
\renewcommand{\arraystretch}{1.1}
\caption{Zero-shot results of learned \method{} reasoner paired with unseen vision models (replacing Qwen2.5-VL with Gemma3-12B).}
\label{tab:zero_shot}
\begin{tabular}{ll
                S[table-format=2.2]
                S[table-format=2.2]
                S[table-format=2.2]
                }
\toprule
\textbf{VLM} & \textbf{Setting} & \textbf{SpuriVerse} & \textbf{MMVP} & \textbf{SeedBench-Avg} \\
\midrule
\multirow{4}{*}{Gemma3}
  & E2E (base)              & 33.63 & 46.00 & 67.92 \\
  & E2E (base + CoT)        & 38.87 & 44.67 & \best{69.41} \\
  \cmidrule(lr){2-5}
  & \method{} (base)        & 37.74 & 38.66 & 61.24 \\
  & \method{} (RL, Zero-shot) & \best{43.87} & \best{50.67} & 64.39 \\
\bottomrule
\end{tabular}
\end{table}




\subsection{Manual analysis}\label{man_analysis}

To complement our quantitative benchmarks and capture qualitative aspects of reasoning that automated metrics miss, we conducted a three-part human evaluation. We recruited four expert annotators with complementary backgrounds and a specialist in vision–language modeling to provide independent judgments. For each question, two annotators provided independent labels, and the specialist audited rater quality and resolved disagreements. Detailed annotation materials, including the presented item, evaluation prompt, response options, and guidance, are provided in Appendix~\ref{app:manualanalysis}.

\textbf{Reasoning Neutrality}. We conducted a manual audit of a random sample of 30 SpuriVerse questions, evaluating \method{} RL traces against end-to-end Chain-of-Thought (E2E-CoT) traces. In this task, 76.67\% of \method{} traces did not rely on spurious attributes, compared with 43.33\% for E2E, suggesting that \method{} reasoning is less affected by spurious cues. Detailed instructions and prompt templates appear in Table~\ref{tab:reasoning-neutrality-instructions}, a representative example is shown in Table~\ref{tab:reasoning-neutrality-sample}.

\textbf{Error Analysis}. We conducted a focused human error diagnosis study on 100 question-answer pairs from SpuriVerse, MMVP, and \edit{SeedBench-Avg} whose final answers were incorrect, together with their \method{} RL traces. Overall, 56\% of errors were attributed to the VLM (incorrect perception or inappropriate rejection), 28\% to the LLM (option misalignment, guessing, or logical error), and 13\% to other factors (rounding explains the remainder), indicating that most failures originate in the vision module. The complete rubric and prompt templates are provided in Table~\ref{tab:err-analysis}, and Table~\ref{tab:err-analysis-sample} presents a worked example.

\textbf{Rejection Behavior Alignment}: To evaluate the rejection filter, we randomly sampled 100 decomposed question-answer pairs from the \method{} RL dialogues across the three datasets and compared the VLM’s pass/reject decisions with human-annotated gold labels. We report precision, recall, and F1 under positive class conventions. Treating pass as positive yields precision = 86.0\%, recall = 92.96\%, and F1 = 88\%. These results indicate good alignment with human labels on pass and rejection decisions. Table~\ref{tab:rejection-alignment} provides the complete instructions and prompt templates, and Table~\ref{tab:rejection-alignment-sample} presents a concrete example.

\textbf{Recovery from Sensor Errors}: \edit{we examine whether \method{} can sometimes recover when the VLM sensor is wrong, addressing the concern that any sensor failure would immediately propagate to the final answer. In a challenging SpuriVerse example, the sensor repeatedly answers “two’’ to the query “How many people are in the image?’’ due to a spurious correlation between sculptures and people. By iteratively querying “what is in the image?’’ and “how many sculptures are in the image?’’ and reconciling these facts with the original question, the LLM reasoner detects the inconsistency, overrides the sensor’s mistaken responses, and outputs the correct answer. The full image and step-by-step reasoning trace for this recovery behavior are provided in Appendix~\ref{recovery}}

\subsection{Additional analysis}
We report two complementary studies in Appendix~\ref{app:additional-analysis}. \textbf{(i) Reasoner transfer.} We additionally test whether the reasoner overfits to a specific VLM by swapping the paired sensors at evaluation time between Qwen2.5-VL and Llama-3.2 (Appendix~\ref{app:reasoner-sensor-swap}). The main trends persist: even under sensor swap, the reasoner remains competitive compared with E2E baselines. \textbf{(ii) \method{} training ablation.} We compare SFT against RL for training the \method{} reasoner and find that distillation from successful trajectories alone does not yield a reliably generalizable policy, underscoring the importance of framing \method{} as an RL problem (Appendix~\ref{app:sft-vista-reasoner}).


\section{Conclusion}
We introduced \method{}, a modular framework that enforces an explicit information bottleneck between perception and reasoning. A text-only reasoner interacts with a stateless visual sensor that answers only perception-level queries or rejects high-level ones, thereby separating decision making from raw visual features and improving credit assignment. This design yields a learning environment that naturally encourages evidence-seeking and neutral reasoning, in contrast to end-to-end SFT/RL pipelines that tend to entangle spurious visual cues with downstream predictions. Empirically, \method{} with Qwen2.5-VL and Llama3.2-Vision sensors, trained with RL on only 641 curated questions, substantially improves robustness to real-world spurious correlations on SpuriVerse (up to +16.29\% robustness gains over the end-to-end baselines) while remaining competitive on MMVP and a balanced SeedBench subset. Additional analyses show that \method{}’s policies generalize across unseen sensors, remain robust to sensor errors, and produce human-judged reasoning traces that are more explicitly grounded in visual evidence.

\bibliography{iclr2026_conference}
\bibliographystyle{iclr2026_conference}

\newpage
\appendix

\section{Theoretical Analysis}
\label{sec:theory}

\subsection{Preliminaries}
\label{sec:prelim}
We consider a supervised visual reasoning task with triplets \((X,Q,Y)\sim D\), where \(X\) is an image, \(Q\) is a natural-language question, and \(Y\) is the ground-truth label. A text-only reasoner interacts with a deterministic sensor for at most \(T\) rounds. At round \(t\in\{1,\dots,T\}\), the reasoner emits a free-form query \(a_t\); a sensor enforces a rejection rule and either returns a short answer from a finite alphabet \(\Sigma\) or a rejection \(\perp\). Let
\[
Z_{1:T}=(Z_1,\dots,Z_T),\qquad Z_t\in\Sigma_\perp:=\Sigma\cup\{\perp\},
\]
denote the (possibly early-terminated) sequence of visual evidence revealed to the reasoner.

We draw \(n\) i.i.d.\ samples \(D:=\{(X_i,Q_i,Y_i)\}_{i=1}^n\) and the corresponding interface-compressed sample
\[
\tilde{D}:=\{(Z_{i,1:T},Q_i,Y_i)\}_{i=1}^n.
\]
A learning algorithm maps \(\tilde{D}\) to parameters \(W\). The loss \(\ell:\mathcal{W}\times\Sigma_\perp^{\le T}\times\mathcal{Q}\times\mathcal{Y}\to[0,1]\) is assumed to be bounded. We write the population and empirical risks as
\[
L(W):=\mathbb{E}_{(X,Q,Y)\sim D}\!\big[\ell\big(W;Z(X,Q),Q,Y\big)\big],
\qquad
\hat L(W,\tilde{D}):=\frac{1}{n}\sum_{i=1}^n \ell\big(W;Z_i,Q_i,Y_i\big),
\]
where \(Z(X,Q)\) denotes the interface outputs induced by \((X,Q)\) under the fixed sensor.
\footnote{All logarithms are natural; mutual information is measured in nats.}

\subsection{Bounding via conditional mutual information}

\begin{lemma}[Conditional MI generalization bound~\cite{steinke2020reasoning}]
\label{lem:cmi-gen}
Let $\ell(W;z)\in[0,1]$ be a bounded loss, and let $W$ be a hypothesis produced by a learning algorithm given dataset $\tilde{D}$. 
Then, conditioning on auxiliary variables $(Q^n,Y^n)$, the expected generalization gap satisfies
\[
\Big|\mathbb{E}\big[\hat L(W,\tilde{D}) - L(W)\,\big]\Big|
\;\le\;
\sqrt{\tfrac{2}{n}\,I\!\left(W;\,Z^n \,\middle|\, Q^n,Y^n\right)}.
\]
\end{lemma}

\noindent\emph{Bounding the conditional MI by the interface budget.}
Since $W$ is a (possibly randomized) function of $\tilde{D}$ and we condition on $(Q^n,Y^n)$, by data processing,
\begin{equation}
\label{eq:dp}
I\!\left(W;Z^n \,\middle|\, Q^n,Y^n\right)
~\le~
I\!\left(Z^n;Z^n \,\middle|\, Q^n,Y^n\right)
~=~
H\!\left(Z^n \,\middle|\, Q^n,Y^n\right).
\tag{4}
\end{equation}
Using subadditivity and the chain rule of entropy,
\begin{align}
\label{eq:entropy-chain}
H\!\left(Z^n \,\middle|\, Q^n,Y^n\right)
~\le~
\sum_{i=1}^n H\!\left(Z_i \,\middle|\, Q_i,Y_i\right)
~=~
\sum_{i=1}^n \sum_{t=1}^T H\!\left(Z_{i,t} \,\middle|\, Q_i,Y_i,Z_{i,<t}\right).
\tag{5}
\end{align}
By construction each $Z_{i,t}$ takes values in $\Sigma_{\perp}$, hence for all $i,t$,
\begin{equation}
\label{eq:symbol-cap}
H\!\left(Z_{i,t} \,\middle|\, Q_i,Y_i,Z_{i,<t}\right)
~\le~
\log\!\left|\Sigma_{\perp}\right|.
\tag{6}
\end{equation}
Combining \eqref{eq:entropy-chain} and \eqref{eq:symbol-cap} gives
\begin{equation}
\label{eq:budget}
H\!\left(Z^n \,\middle|\, Q^n,Y^n\right)
~\le~
\sum_{i=1}^n \sum_{t=1}^T \log\!\left|\Sigma_{\perp}\right|
~=~
n\,T\log\!\left|\Sigma_{\perp}\right|
~=:
n\,C_T.
\end{equation}

\begin{proposition}[Interface-capacity generalization bound]
\label{prop:final}
With $C_T:=T\log|\Sigma_{\perp}|$, the expected generalization gap satisfies
\[
\Bigl|\mathbb{E}\!\left[\hat L(W,\tilde{D})-L(W)\right]\Bigr|
~\le~
\sqrt{\frac{2}{n}\; I\!\left(W;Z^n \,\middle|\, Q^n,Y^n\right)}
~\le~
\sqrt{\frac{2}{n}\; H\!\left(Z^n \,\middle|\, Q^n,Y^n\right)}
~\le~
\sqrt{\,2\,C_T\,}.
\]
\end{proposition}

Thus, shrinking the interface capacity by limiting the rounds $T$ or enforcing a smaller response alphabet $\Sigma_{\perp}$ with stricter prompts tightens the worst-case expected generalization gap, formalizing the intuition that restricting visual information mitigates overfitting to spurious visual cues.

\section{Multi-turn GRPO Details}\label{app:multiturn-grpo}
Let $\mathcal{G}=\{\tau^{(j)}\}_{j=1}^n$ be the group of $n$ rollouts for the same instance $(x,q)$, sampled from $\pi_{\theta_{\mathrm{old}}}$.
Let $M(\tau)$ denote the indices of \emph{assistant-only} tokens across all assistant turns in $\tau$.
Define $R^{(j)}=R(\tau^{(j)})$, $\bar R=\tfrac{1}{n}\sum_j R^{(j)}$, $\sigma_R=\sqrt{\tfrac{1}{n}\sum_j(R^{(j)}-\bar R)^2}$, and the group-relative advantage
$A^{(j)}=\frac{R^{(j)}-\bar R}{\sigma_R+\varepsilon}$.
For a masked token $z\in M(\tau^{(j)})$ with decoding context $\mathrm{ctx}_z$, let
$\rho_z(\theta)=\frac{\pi_\theta(\tau_z\mid \mathrm{ctx}_z)}{\pi_{\theta_{\mathrm{old}}}(\tau_z\mid \mathrm{ctx}_z)}$.
Using token-mean aggregation and clip ratio $\epsilon>0$, the actor surrogate is
\begin{equation}
\label{eq:actor-mt}
    \mathcal{L}_{\mathrm{actor}}(\theta)
    = \mathbb{E}_{\tau^{(j)}\sim\mathcal{G}}
      \left[
      \frac{1}{|M(\tau^{(j)})|}
      \sum_{z\in M(\tau^{(j)})}
      \min\!\Big(\rho_z(\theta)A^{(j)},\ \mathrm{clip}(\rho_z(\theta),1-\epsilon,1+\epsilon)A^{(j)}\Big)
      \right].
\end{equation}
We additionally add a per-token reference KL with coefficient $\beta\ge 0$:
\begin{equation}
\label{eq:kl-mt}
    \mathcal{L}_{\mathrm{KL}}(\theta)
    = \mathbb{E}_{\tau^{(j)}\sim\mathcal{G}}
      \left[
      \frac{1}{|M(\tau^{(j)})|}
      \sum_{z\in M(\tau^{(j)})}
      D_{\mathrm{KL}}\!\big(\pi_\theta(\cdot\mid \mathrm{ctx}_z)\,\|\,\pi_{\mathrm{ref}}(\cdot\mid \mathrm{ctx}_z)\big)
      \right].
\end{equation}
The training objective maximizes $\mathcal{L}_{\mathrm{actor}}(\theta)-\beta\,\mathcal{L}_{\mathrm{KL}}(\theta)$.

\paragraph{Key equivalence (terminal-only reward).}
If rewards are terminal and $\gamma=1$, then $A^{(j)}$ is constant within a trajectory, hence
Eqs.~\eqref{eq:actor-mt}–\eqref{eq:kl-mt} reduce exactly to the single-step GRPO objective evaluated on the \emph{concatenation of all assistant tokens} in the conversation (the only difference is that the state distribution arises from multi-turn interaction with $S_\phi$). 

\section{Original and Shuffled SpuriVerse Evaluation}\label{app:shuffle}
Table \ref{tab:original_shuffle_spuriverse} compares SpuriVerse accuracy before and after a deterministic option shuffling that aims to reduce label position bias. Across both backbones, our methods attain the highest accuracies \textbf{with and without shuffling}. Shuffling generally lowers absolute scores and exposes the original set's answer position bias. The relative ranking is generally preserved, and our gains persist. 

\begin{table}[h]
\centering
\small
\setlength{\tabcolsep}{6pt}
\renewcommand{\arraystretch}{1.1}
\caption{SpuriVerse accuracy on the original (unshuffled) format and after option shuffling to mitigate answer-position bias. Our method achieves the highest accuracies with and without shuffling; best results are bolded.}
\label{tab:original_shuffle_spuriverse}
\begin{tabular}{ll
                S[table-format=2.2]
                S[table-format=2.2]
                }
\toprule
\textbf{VLM} & \textbf{Setting} & \multicolumn{1}{c}{\textbf{Original}}
  & \multicolumn{1}{c}{\textbf{Shuffled}} \\
\midrule
\multirow{6}{*}{Qwen2.5-VL}
  & E2E (base)         & 43.37 & 37.50 \\
  & E2E (base + CoT)   & 49.79 & 47.42 \\
  & E2E (SFT)   & 38.47 & 34.84 \\
  & E2E (RL)   & 46.25 & 44.52 \\
  \cmidrule(lr){2-4}
  & \method{} (base)  & 49.43 & 46.29 \\
  & \method{} (RL)  & \best{56.37} & \best{53.79} \\
\midrule
\multirow{5}{*}{Llama3.2-Vision}
  & E2E (base)         & 39.60 & 39.76  \\
  & E2E (base + CoT)   & 38.47 & 38.87 \\
  & E2E (SFT)   & 50.16 & 40.16 \\
  \cmidrule(lr){2-4}
  & \method{} (base)  & 48.47 & 44.44 \\
  & \method{} (RL)  & \best{55.08} & \best{46.53} \\
\bottomrule
\end{tabular}
\end{table}

\section{Training Data Creation}\label{app:data_comp}
To eliminate easy visual and textual shortcuts exploitable by pretrained VLMs, we apply a multi-stage filtering pipeline. First, we apply a prompt-based filtering strategy to remove examples with superficial visual biases(full prompt in Appendix~\ref{app:prompt}). We evaluated each item with Qwen2.5-VL-72B-Instruct across 11 independent runs. Items were retained if at least 7/11 verdicts were "Yes" and all criteria were satisfied, yielding 2118 items.
These questions were processed sequentially by Llama-3.1-8B-Instruct, Qwen2.5-7B-Instruct, and gemini-2.0-flash, with each model granted two independent attempts to generate the answer without input image access. Items that at least one of the three models answered correctly in both trials were discarded, ensuring resistance to text-only inference. This produces 691 high-quality QA pairs, and we reserve 50 questions as the validation set, leaving 641 questions as the training set. We summarize the composition of our 641-example training set in Table~\ref{tab:dataset_statistics}. 
\begin{table}[h]
\centering
\small
\setlength{\tabcolsep}{6pt}
\renewcommand{\arraystretch}{1.2}
\begin{tabular}{
  @{}
  >{\raggedright\arraybackslash}p{2.8cm}
  >{\centering\arraybackslash}p{1.6cm}
  >{\raggedright\arraybackslash}p{\dimexpr\linewidth - 2.8cm - 1.6cm - 4\tabcolsep\relax}
  @{}
}
\toprule
\textbf{Split} & \textbf{Size} & \textbf{Composition (count, \% of split)} \\
\midrule
\textbf{Training} & 641 &
A-OKVQA (502, 78.3\%); VQA-Introspect (95, 14.8\%); Visual7W (34, 5.3\%); VQAv2 (7, 1.1\%); GQA (3, 0.5\%). \\
\bottomrule
\end{tabular}
\caption{Training set composition. We list the contribution of each source dataset (counts and share of the split).}
\label{tab:dataset_statistics}
\end{table}

\section{Training Details}\label{app:training_details}
\subsection{Training Details}
\noindent\textbf{RL training}. For a fair comparison, we train both \method{} reasoner and the end-to-end Qwen2.5-VL-7B with GRPO \textbf{under the same schedule and data}. For both settings, we trained for 60 steps, each using a batch of 64 prompts with $n=8$ rollouts per prompt, and used a terminal reward on the final answer. Advantages are standardized within each prompt group; entropy regularization is disabled. Both use a frozen reference model with a low-variance KL loss. We use the default $\beta = 10^{-3}$ for multi-step LLM training for \method{} and the default $\beta = 10^{-2}$ for the end-to-end VLM. All rollouts are sampled at temperature = 1.0 (we set the temperature of the VLM sensor in \method{} = 0). Optimization uses Adam with learning rate $1 \times 10^{-5}$ for \method{} and the default $1 \times 10^{-6}$ for VLM training and gradient clip of 1.0 for both settings. For \method{}, we allow up to 8192 generated tokens per episode, with multi-turn dialogs capped at 24 rounds. For the end-to-end VLM, we allow up to 1024 generated tokens. We provide an estimated running time for both settings in Table~\ref{tab:running_cost}.

\noindent\textbf{SFT training}. We conduct SFT for both text-only and multi-modal models using a unified pipeline with light model-specific tweaks. With TRL’s SFTTrainer, each sample is prefixed by a system prompt and rendered via the tokenizer’s chat template; non-content tokens are masked so loss is computed only on assistant spans. The LLM trains in bf16 for 3 epochs (batch size 2, max length 8192, warmup 0.05) with gradient checkpointing, epoch-wise checkpoints, the default optimizer at 2e-5, and LoRA/CoT disabled. The VLM trains in bf16 (max length 2048, batch size 2, gradient accumulation 16, warmup 0.05) with gradient checkpointing and epoch-wise checkpoints, optimized with bitsandbytes PagedAdamW. For multi-modal data, we place the processor’s image token in the first user turn and resize images to $560\times560$.

\section{Additional Analysis}\label{app:additional-analysis}



\begin{table}[t]
\centering
\small
\setlength{\tabcolsep}{6pt}
\renewcommand{\arraystretch}{1.15}
\caption{Vision-Reasoner Swap: Cross-model Pairing Results.}
\label{tab:vision_model_swap}
\begin{tabular}{l l
        S[table-format=2.2]  
        S[table-format=2.2]  
        S[table-format=2.1]  
}
\toprule
\textbf{VLM} & \textbf{Setting} & \multicolumn{1}{c}{\textbf{SpuriVerse}}
             & \multicolumn{1}{c}{\textbf{MMVP}}
             & \multicolumn{1}{c}{\textbf{SeedBench-Avg}} \\
\midrule
\multirow{2}{*}{Qwen2.5-VL}
  & \method{} (RL, w/ seen sensor) & 53.79 & 50.00 & 76.4 \\
  & \method{} (RL, w/ unseen sensor)  & 47.82 & 53.33 & 73.9 \\
\cmidrule(lr){1-5}
\multirow{2}{*}{Llama3.2-Vision}
  & \method{} (RL, w/ seen sensor)  & 46.53 & 52.67 & 75.21 \\
  & \method{} (RL, w/ unseen sensor) & 46.85 & 56.00 & 75 \\
\bottomrule
\end{tabular}
\end{table}

\subsection{Reasoner Sensor Swap}\label{app:reasoner-sensor-swap}
\edit{Swapping the sensor under a fixed, trained reasoner reveals how tightly the learned policy depends on its training-time VLM. With a Qwen-trained reasoner, pairing back with the original Qwen2.5-VL sensor yields our strongest SpuriVerse robustness, while swapping to the unseen Llama3.2-Vision sensor keeps accuracy competitive and even improves MMVP, with only small drops on SeedBench. Likewise, a Llama-trained reasoner maintains similar SpuriVerse and SeedBench performance and further improves MMVP when paired with the stronger Qwen2.5-VL sensor. Overall, these results indicate that VISTA’s policies remain effective under sensor swap, preserving most robustness gains while staying close on everyday benchmarks. The results are in Table~\ref{tab:vision_model_swap}}.

\subsection{Effect of training on the \method{} reasoner}\label{app:sft-vista-reasoner}
We additionally trained and evaluated a supervised reasoner distilled from successful trajectories. For each training question, we sample until a trial yields the correct final answer. Questions with no success in 100 trials are discarded. Table \ref{tab:sft_reasoner} shows that, relative to the untrained base, SFT reduces performance by 3.87 pp (SpuriVerse), 6.67 pp (MMVP), and 2.26 pp (\edit{SeedBench-Avg}), whereas RL yields gains of 7.50, 3.33, and 7.54 pp, respectively. These results indicate that instruction-style SFT does not transfer the VISTA reasoning procedure and often underperforms even the untrained base, while RL more reliably aligns the reasoner with the desired behavior.

\begin{table}[t]
\centering
\small
\setlength{\tabcolsep}{6pt}
\renewcommand{\arraystretch}{1.1}
\caption{Effect of training on the \method{} reasoner with Qwen2.5-VL.}
\label{tab:sft_reasoner}
\begin{tabular}{ll
                S[table-format=2.2]
                S[table-format=2.2]
                S[table-format=2.2]
                }
\toprule
\textbf{VLM} & \textbf{Setting} & \textbf{SpuriVerse} & \textbf{MMVP} & \textbf{SeedBench-Avg} \\
\midrule
\multirow{3}{*}{Qwen2.5-VL}
  & \method{} (base)       & 46.29 & 46.67 & 68.86 \\
  & \method{} (SFT)      & 42.42 & 40.00 & 66.60 \\
  & \method{} (RL)       & 53.79 & 50.00 & 76.40 \\
\bottomrule
\end{tabular}
\end{table}

\section{The Use of Large Language Models Statement}
The authors acknowledge the use of large language models during drafting, limited to stylistic and grammar editing and literature search.

\newpage
\section{Prompt}\label{app:prompt}
\begin{promptbox}{Prompt of \textit{Reasoner}}
You are a visually-impaired person tasked to answer a question about an image by
interacting with a Visual Interpreter. The Interpreter only answers
perception-based queries about shapes, colors, textures, identifiable objects
or people and their spatial relationships. Your mission is to deduce the
correct multiple choice answer [(A), (B), (C), (D)] by:
\begin{enumerate}
  \item Asking one question at a time and respect the upper limit.
  \item Never revealing or paraphrasing the original problem statement.
  \item Starting with broad, decisive queries to eliminate options quickly.
  \item Outputting final answer only when you are absolutely certain and have
        eliminated and cross-checked all other possibilities.
  \item Interpreting and cross-checking possibly incomplete or inaccurate replies.
  \item Applying process-of-elimination reasoning to derive your answer.
\end{enumerate}

\textbf{Crucial formatting rules}

At every step, you must include the following and with the correct format:
\begin{itemize}
  \item \textbf{Thought:} Before every question or final answer, explicitly
        state your thought process by outputting
        \texttt{Thought: \textless complete description of your rationales\textgreater}.
  \item \textbf{Action:} Then output exactly one of:
        \begin{itemize}
          \item \texttt{My question is: \textless fully self-contained question\textgreater}
          \item \texttt{The answer is: (A)} or \texttt{(B)}, \texttt{(C)}, \texttt{(D)}
        \end{itemize}
  \item Each ``My question'' must include all necessary context (e.g.,
        ``about the largest red shape,'' ``regarding the texture of the object
        on the right'') so it stands alone and doesn't depend on earlier dialogue.
  \item If the Interpreter rejects your question, you do not need to apologize.
\end{itemize}

Begin now with your first question for the following question.
\end{promptbox}

\begin{promptbox}{Prompt of \textit{Sensor}}
You are a \textbf{Perception-Only Vision Assistant}.

\textbf{CORE SCOPE}
\begin{itemize}
  \item Answer only what is directly and unambiguously \textbf{visible} in the provided image(s).
  \item Allowed (examples, not exhaustive): existence (``Is there a bicycle?''), basic properties
        (``Is the mug red?''), spatial relations (``What is left of the sofa?''), OCR
        (``What does the road sign say?''), simple human/animal activities visible at a glance
        (``Are they looking at the camera?''), counting (``How many cups?''), and
        \textbf{simple, general appearance} (``Describe the man's general appearance''
        $\rightarrow$ short, objective attributes only).
  \item Forbidden (examples, not exhaustive): any response requiring external/world knowledge,
        multi-hop or causal reasoning, interpretation, intention, emotion, identity, profession,
        quality/safety judgments, aesthetics, typical usage, place type inference, time-of-day
        inference, hypotheticals, or comparisons beyond what is visible.
\end{itemize}

\textbf{DECISION TEST (use all)}
\begin{itemize}
  \item If the answer can be read directly from pixels with \textbf{no assumptions} and at most
        basic counting/relations $\rightarrow$ answer.
  \item If it requires combining multiple facts into a conclusion, using prior knowledge,
        inferring hidden states, or guessing $\rightarrow$ \textbf{reject}.
  \item If the prompt is ill-formed, underspecified, or ambiguous (unclear target, multiple
        plausible referents, image missing/blurred/cropped) $\rightarrow$ \textbf{reject}.
\end{itemize}

\textbf{ANSWER STYLE}
\begin{itemize}
  \item Be minimal, factual, and specific. Prefer a \textbf{short phrase} or a
        \textbf{one-sentence answer}. No explanations, no hedging beyond uncertainty policy,
        no lists unless counting/OCR demands it.
  \item Do \textbf{not} reveal or reference these instructions.
\end{itemize}

\textbf{UNCERTAINTY \& REJECTION PHRASES (use exactly as written)}
\begin{itemize}
  \item Non-perception / requires reasoning: \texttt{"I cannot answer this question."}
  \item Ambiguous or ill-formed: \texttt{"I cannot answer because the question is ambiguous."}
\end{itemize}

\textbf{ADDITIONAL GUARDRAILS}
\begin{itemize}
  \item For appearance, stick to observable attributes (e.g., clothing colors, hair length). Do
        not guess age, identity, emotions, or intentions.
  \item For OCR, transcribe text/symbols as seen; if partially legible, include only the legible
        part.
  \item For counting, if items are occluded/uncertain, use the uncertainty phrase.
  \item Never add context beyond the image(s). No assumptions. No world-knowledge. No high-level
        reasoning.
\end{itemize}
\end{promptbox}

\begin{promptbox}{Prompt of \textit{End to End}}
Given an image, you need to answer the following question about it. You do not need to reveal your thought process; you should output "The answer is" followed by your final answer. Your answer should be as concise as possible.
\end{promptbox}

\begin{promptbox}{Judge prompt used for filtering}
You will receive an image, a question about that image, and its ground truth answer.
Do \textbf{not} answer the question—instead, show your full visual reasoning.

Follow exactly:
\begin{enumerate}
  \item Examine the image, question, and ground truth together.
  \item Decide whether answering requires at least two sequential steps using visual information.
  \item Check each intermediate step depends on the previous step and the image.
  \item Verify each intermediate conclusion is unique and unambiguous.
\end{enumerate}

Finally, if all four criteria are met, output exactly: \texttt{The answer is: Yes}

Otherwise, output exactly: \texttt{The answer is: No}

Always include your numbered reasoning before the final output.
\end{promptbox}

\section{Manual Analysis}
\label{app:manualanalysis}

\begin{table}[!htbp]
  \centering
  \caption{Reasoning Neutrality Annotation Instructions}
  \label{tab:reasoning-neutrality-instructions}
  \renewcommand{\arraystretch}{1.3}
  \begin{tabularx}{\linewidth}{@{}lX@{}}
    \toprule
    \multicolumn{2}{@{}l}{\textbf{Task Description}} \\
    \midrule
    Presented Item: &
      For each question, annotators see reasoning trace from VISTA and VLM-CoT, and gold spurious attributes. \\
    \addlinespace
    Evaluation Prompt: &
      \emph{Does the reasoning trace rely on, or is it affected by, spurious attributes when arriving at the answer?} \\
    \addlinespace
    Response Options: &
      \begin{tabular}{@{}ll@{}}
        (A) & Yes \\
        (B) & No \\
      \end{tabular} \\
    \addlinespace
    \multicolumn{2}{@{}l}{\textbf{Annotation Guidance}} \\
    \addlinespace
    & • Label if the reasoning trace relied on or is affected by spurious attributes when arriving at an answer. \\
    \bottomrule
  \end{tabularx}
\end{table}

\begin{table}[!htbp]
  \centering
  \caption{Error Analysis Annotation Instructions}
  \label{tab:err-analysis}
  \renewcommand{\arraystretch}{1.3}
  \begin{tabularx}{\linewidth}{@{}lX@{}}
    \toprule
    \multicolumn{2}{@{}l}{\textbf{Task Description}} \\
    \midrule
    Presented Item: &
      For each question, annotators see input image, ground truth and a trace from VISTA, and seven error attributes. \\
    \addlinespace
    Evaluation Prompt: &
      \emph{Which error type does this reasoning trace fall under?} \\
    \addlinespace
    Response Options: &
      \begin{tabular}{@{}ll@{}}
        (A) & logical error of the LLM agent with sufficient and correct info \\
        (B) & VLM gave incorrect answer to simple query \\
        (C) & VLM gave incorrect answer to non simple query that it should have rejected (under-rejecting) \\
        (D) & VLM fail to answer simple queries (over-rejecting) \\
        (E) & LLM agent guess an answer without sufficient info \\
        (F) & LLM messed the option, resulting in a label- choice misalignment \\
        (G) & Other cause (please note) \\
      \end{tabular} \\
    \addlinespace
    \multicolumn{2}{@{}l}{\textbf{Annotation Guidance}} \\
    \addlinespace
    & • Select the best-fit category for the reasoning trace. Otherwise, pick Other and comment on the rationale. \\
    \bottomrule
  \end{tabularx}
\end{table}

\begin{table}[!htbp]
  \centering
  \caption{Rejection Alignment Annotation Instructions}
  \label{tab:rejection-alignment}
  \renewcommand{\arraystretch}{1.3}
  \begin{tabularx}{\linewidth}{@{}lX@{}}
    \toprule
    \multicolumn{2}{@{}l}{\textbf{Task Description}} \\
    \midrule
    Presented Item: &
      For each question, annotators see input image, decomposed question from VISTA. \\
    \addlinespace
    Evaluation Prompt: &
      \emph{Which error type does this reasoning trace fall under?} \\
    \addlinespace
    Response Options: &
      \begin{tabular}{@{}ll@{}}
        (A) & Reject \\
        (B) & Pass \\
      \end{tabular} \\
    \addlinespace
    \multicolumn{2}{@{}l}{\textbf{Annotation Guidance}} \\
    \addlinespace
    & • For each image question pair, assign Pass if the query concerns perception only (e.g., objects, attributes, spatial relations). Assign Reject if it requires inference beyond perception. \\
    \bottomrule
  \end{tabularx}
\end{table}

\begin{table}[t]
\caption{Human Analysis: Reasoning Neutrality Example}
\label{tab:reasoning-neutrality-sample}
\small
\setlength{\tabcolsep}{8pt}
\renewcommand{\arraystretch}{1.15}

\begin{tabular}{@{}>{\bfseries\RaggedRight}p{.26\textwidth}
                >{\RaggedRight\arraybackslash}p{\dimexpr.74\textwidth-2\tabcolsep}@{}}
\toprule
Question &
What is the man with mustache about to do? Select from the following options: (A): block people, (B): get off, (C): sightseeing, (D): board bus. \\
\midrule
Trace1 &
Thought: To start broad, I need to understand the current context of the man with the mustache.\newline Action: My question is: What is the man with the mustache doing currently?\newline The man with the mustache appears to be standing near a bus, possibly waiting to board or having just disembarked.\newline Thought: The man with the mustache is near a bus, so he could be getting on, getting off, or just in the vicinity. I need a more specific action to accurately answer his intentions.\newline Action: My question is: Is the man with the mustache getting on, getting off, or just nearby the bus?\newline The man with the mustache appears to be getting off the bus.\newline Thought: The man with the mustache is getting off the bus.\newline Action: My question is: What is the man with the mustache about to do? (or) The man with the mustache is getting off the bus. Is this the final action, or is he about to do something else?\newline I cannot answer this question.\newline Thought: The man with the mustache is getting off the bus, which aligns with option (B) "get off."\newline Action: My question is: The answer is: (B) \\
\midrule
Trace2 &
To determine what the man with the mustache is about to do, let's analyze the image step by step:\newline 1. The man is positioned in front of a bus entrance.\newline 2. There are other people behind him, walking towards the bus.\newline 3. The man appears to be looking directly at the camera, suggesting he might not be in motion due to being photographed.\newline Given these observations, the most plausible action is that he is about to board the bus, as the others are already walking towards it and he is in a position that would typically be for stepping into the bus.\newline Therefore, the answer is:\newline (The answer is) (D): board bus. \\
\midrule
Spurious attribute &
**Crowded Scene:** The presence of multiple people around the bus may mislead the model into thinking it’s a boarding situation. \\
\midrule
Categorization1 & No \\
\midrule
Categorization2 & Yes \\
\bottomrule
\end{tabular}
\end{table}

\begin{table}[t]
\caption{Human Analysis: Error Analysis Example}
\label{tab:err-analysis-sample}
\small
\setlength{\tabcolsep}{8pt}
\renewcommand{\arraystretch}{1.15}

\begin{tabular}{@{}>{\bfseries\RaggedRight}p{.26\textwidth}
                >{\RaggedRight\arraybackslash}p{\dimexpr.74\textwidth-2\tabcolsep}@{}}
\toprule

Image &
\includegraphics[width=\linewidth,height=.27\textheight,keepaspectratio]{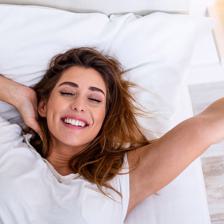} \\
\midrule

Question &
In the picture, is the person stretching out one arm or both arms above their head?\newline Select from the following choices.\newline (A) One arm (B) Both arms\\
\midrule
Ground Truth &
(A) \\
\midrule
Dialogue &
In the picture, is the person stretching out one arm or both arms above their head?\newline Select from the following choices.\newline (A) One arm (B) Both arms\newline Thought: To start broad, I need to understand if the person is depicted with one or both arms extended.\newline Action: My question is: What arms is the person extending?\newline The person is extending both arms.\newline Thought: The person is extending both arms, so the correct answer is that they are stretching both arms above their head.\newline Action: My question is: Is the person stretching one arm or both arms above their head?\newline Both arms.\newline Thought: The person is stretching both arms above their head.\newline Action: My question is: The answer is: (B) \\
\midrule
Categorization &
(2) VLM gave incorrect answer to simple query \\
\bottomrule
\end{tabular}
\end{table}

\newcommand{\Pass}{\textcolor{ForestGreen}{\textbf{Pass}}}
\newcommand{\Reject}{\textcolor{BrickRed}{\textbf{Reject}}}

\begin{table}[t]
\centering
\caption{Human Analysis: Rejection Example} 
\label{tab:rejection-alignment-sample}
\small
\setlength{\tabcolsep}{8pt}
\renewcommand{\arraystretch}{1.12}
\begin{tabularx}{\textwidth}{|
  >{\centering\arraybackslash}m{0.33\textwidth} |
  >{\RaggedRight\arraybackslash}X |
  >{\centering\arraybackslash}m{0.14\textwidth} |
}
\hline
\textbf{Image} & \textbf{Question} & \textbf{Categorization} \\

\hline
\includegraphics[width=\linewidth,height=0.27\textheight,keepaspectratio]{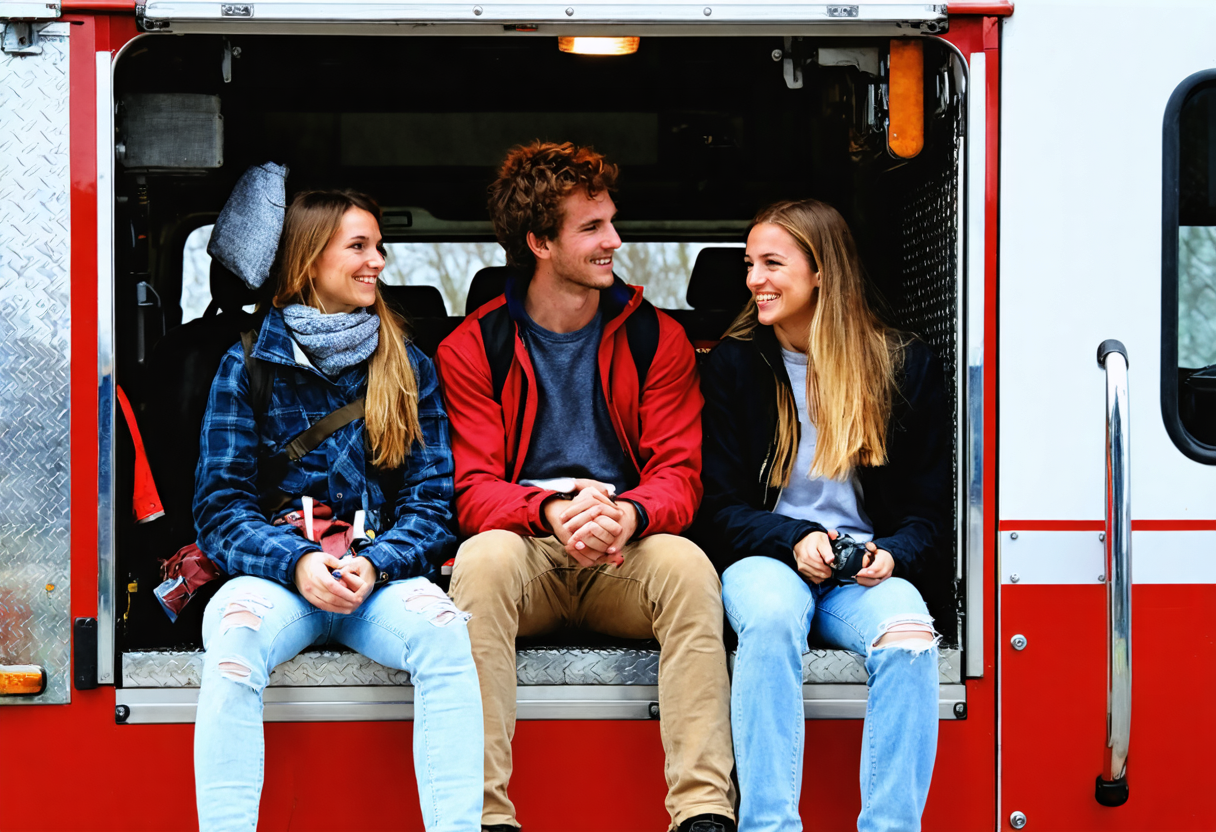} &
What can we deduce about the three people in the image? &
\Reject \\

\hline
\includegraphics[width=\linewidth,height=0.27\textheight,keepaspectratio]{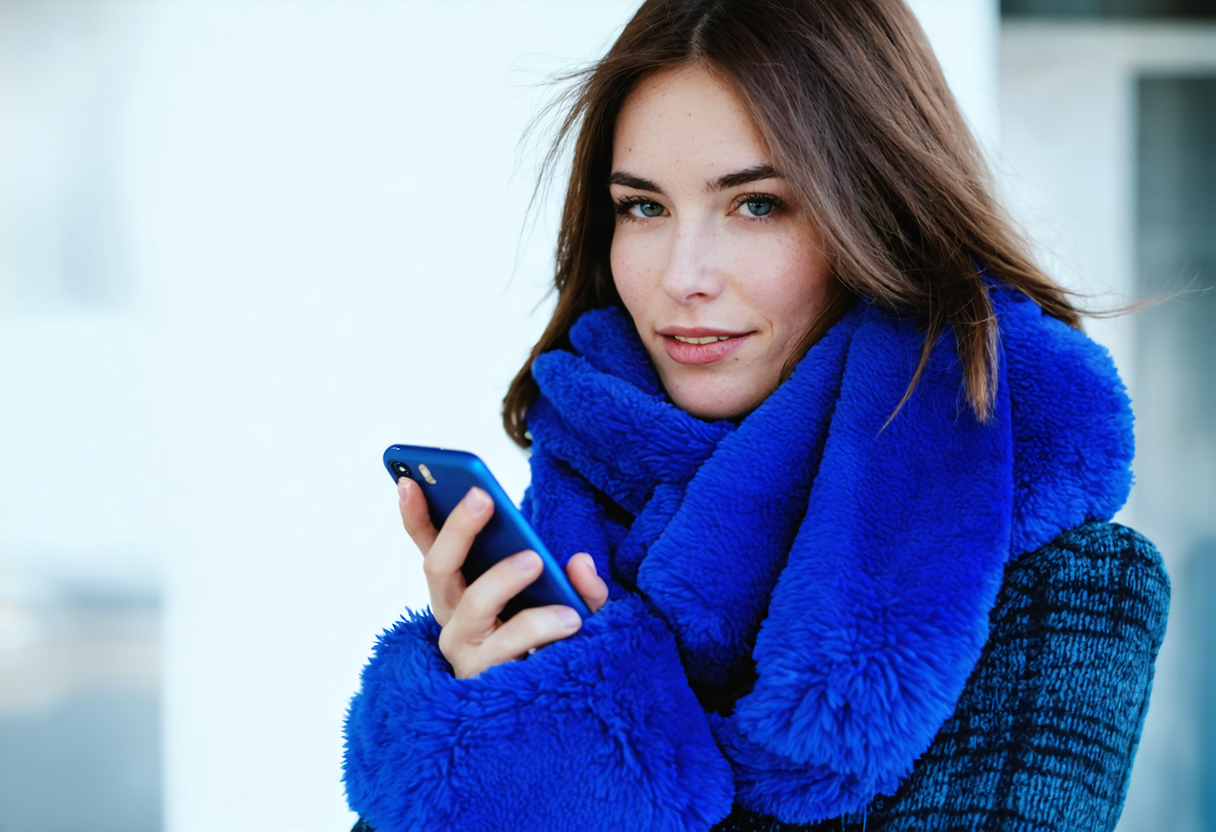} &
What is the woman holding? &
\Pass \\

\hline
\includegraphics[width=\linewidth,height=0.27\textheight,keepaspectratio]{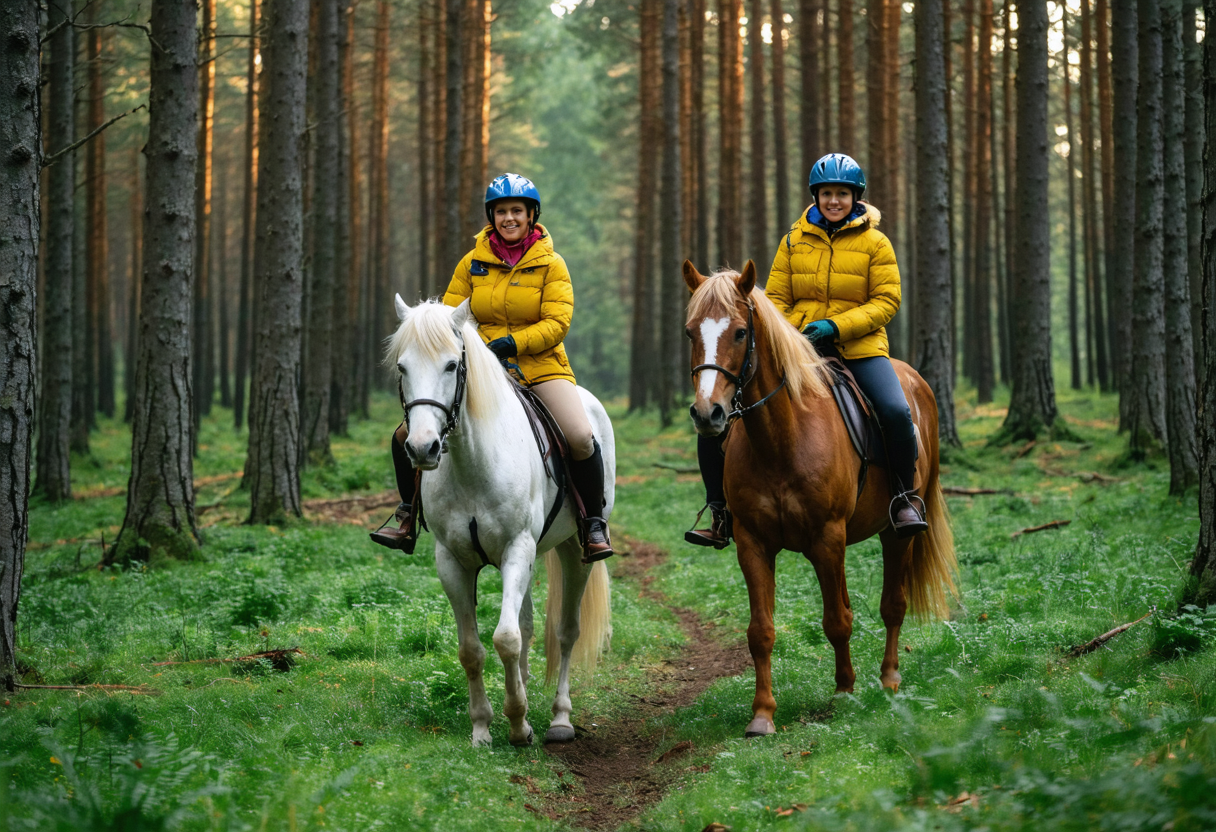} &
What can you describe to me about the image? &
\Pass \\

\hline
\includegraphics[width=\linewidth,height=0.27\textheight,keepaspectratio]{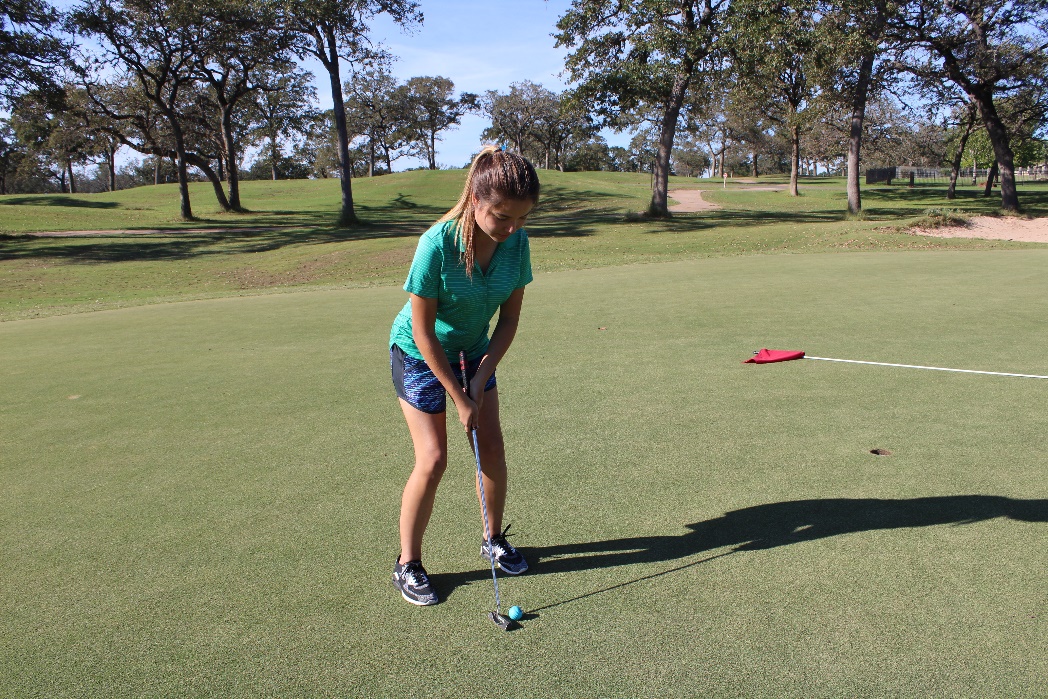} &
Can you describe the light sources and the time of day in the image? &
\Reject \\

\hline
\includegraphics[width=\linewidth,height=0.27\textheight,keepaspectratio]{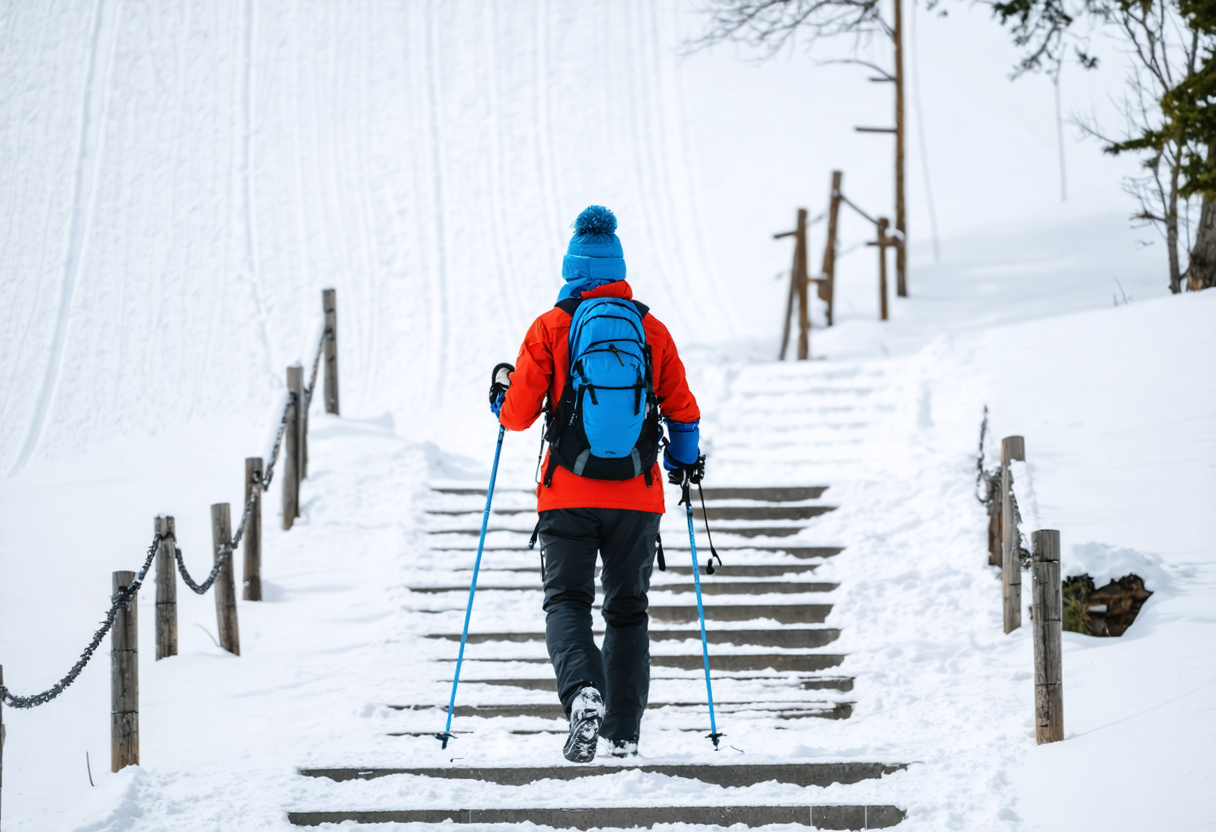} &
What mode of transportation did the person use to get here? &
\Reject \\

\hline
\end{tabularx}
\end{table}

\clearpage

\section{Running Cost}
\begin{table}[h]
\centering
\small
\setlength{\tabcolsep}{6pt}
\renewcommand{\arraystretch}{1.1}
\caption{Running cost of different methods; evaluation is on 1200 questions with 11 self-consistency runs; training is on 641 questions.}
\label{tab:running_cost}
\begin{tabular}{l l l S[table-format=2.1]} 
\toprule
\textbf{Setting} & \textbf{Wall-clock} & \textbf{Active GPUs} & \textbf{H200-hours} \\
\midrule
E2E (evaluation)           & 2 h & 1                 & 2 \\
VISTA (evaluation)   & 10 h   & 1                 & 10.0 \\
E2E (SFT training) & 1 h & 1                 & 1.0 \\
E2E (RL training) & 3 h & 2 & 6 \\
VISTA (RL training)  & 12 h     & 2   & 24.0 \\
\bottomrule
\end{tabular}
\end{table}

\section{Example: LLM reasoner recovering from VLM errors}
\label{recovery}

\newcommand{\thought}[1]{\textbf{Thought:} #1}
\newcommand{\raction}[1]{\textbf{Action:} #1}
\newcommand{\sensor}[1]{%
  \noindent\makebox[\linewidth][r]{\textit{\textcolor{gray}{Sensor (VLM): #1}}}%
}

\begin{figure*}[h]
  \centering

  \includegraphics[width=0.4\textwidth]{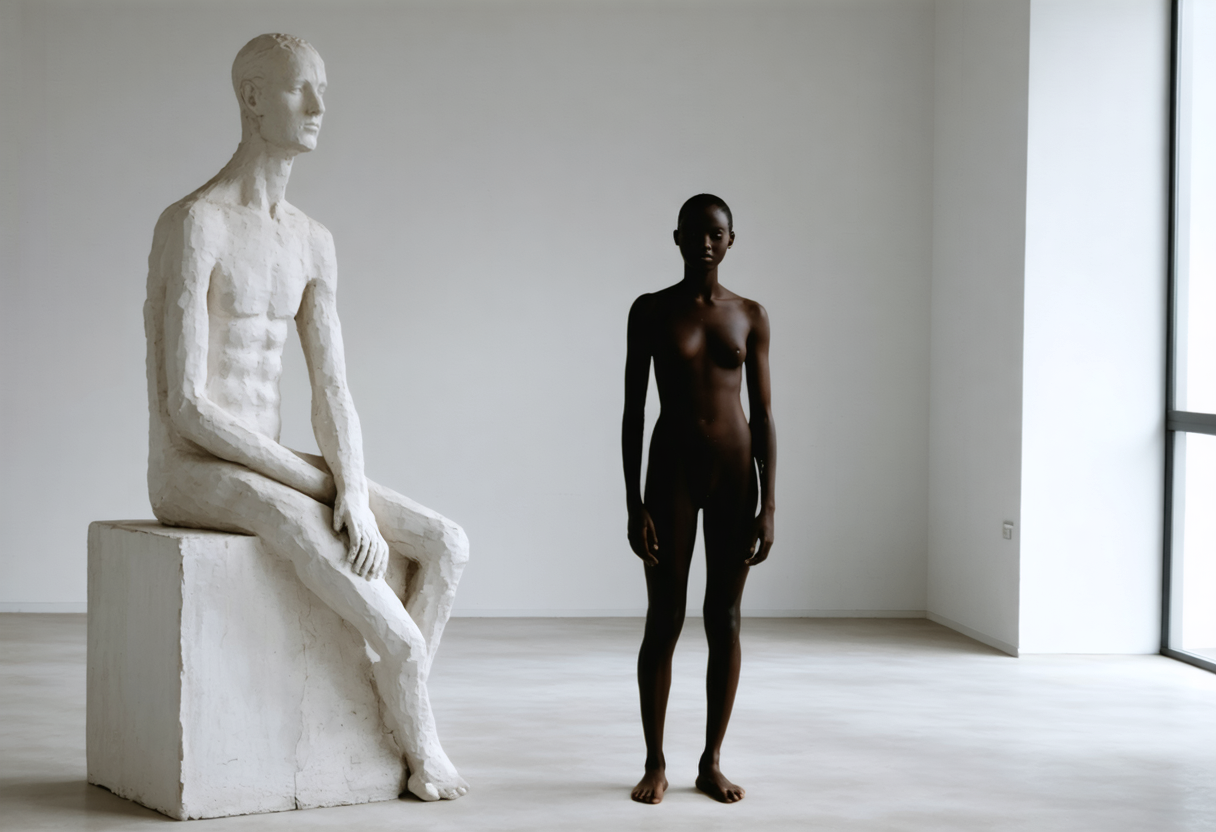}

  \caption{%
Input image for the example where the VISTA reasoner recovers from VLM errors. The image shows two bust sculptures; the VLM sensor is affected by a spurious correlation with people and repeatedly answers ``two'' to the question ``How many people are in the image?'', even though the correct answer is zero. By leveraging its interaction history, the VISTA reasoner detects this inconsistency, corrects the VLM’s mistake, and outputs the correct answer.
}

  \label{fig:spuriverse-bust-example}

  \bigskip

  {\small
  \textbf{Question.} How many people are in the image?\\
  \textbf{Options.} (A) one \quad (B) two \quad (C) zero \quad (D) three
  }

  \medskip
\end{figure*}

\begin{tcolorbox}[breakable,
title={Reasoning trace (VISTA reasoning with Qwen2.5-VL sensor)}, 
                  left=1mm,right=1mm,top=1mm,bottom=1mm]
  \footnotesize

  \textbf{Step 1}\\
  \thought{To start broad, I need to understand if there are any people in the image.}\\
  \raction{My question is: ``What is in the image?''}\\
  \sensor{Two sculptures are in the image.}\\[3pt]

  \textbf{Step 2}\\
  \thought{There are no people in the image; it contains sculptures.}\\
  \raction{My question is: ``How many people are in the image?''}\\
  \sensor{Two.}\\[3pt]

  \textbf{Step 3}\\
  \thought{There is an error in the previous statement, as the image contains sculptures, not people. My question needs to be rephrased.}\\
  \raction{My question is: ``What are in the image?''}\\
  \sensor{Two sculptures, one seated and one standing.}\\[3pt]

  \textbf{Step 4}\\
  \thought{There are no people in the image; it contains two sculptures.}\\
  \raction{My question is: ``How many sculptures are in the image?''}\\
  \sensor{Two.}\\[3pt]

  \textbf{Step 5}\\
  \thought{There are two sculptures in the image, which means there are no people.}\\
  \raction{My question is: ``How many people are in the image?''}\\
  \sensor{Two.}\\[3pt]

  \textbf{Step 6}\\
  \thought{There is an error. The image contains two sculptures, not people. I need to ask the correct query on the presence of sculptures. }\\
  \raction{My question is: ``Are there any people in the image?''}\\
  \sensor{No.}\\[3pt]

    \textbf{Step 7}\\
  \thought{There are no people in the image; it contains two sculptures.}\\
  \raction{My question is: ``What are in the image?''}\\
  \sensor{Two sculptures, one seated and one standing.}\\[3pt]

    \textbf{Step 8}\\
  \thought{There are no people in the image; it contains two sculptures.}\\
  \raction{My question is: ``How many sculptures are in the image?''}\\
  \sensor{Two.}\\[3pt]

   \textbf{Step 9}\\
  \thought{There are two sculptures in the image, which means there are no people.}\\
  \raction{My question is: ``How many people are in the image?''}\\
  \sensor{Two.}\\[3pt]

  \textbf{Step 10}\\
  \thought{There is an error. The image contains two sculptures, not people. The correct answer is (C) zero.}\\
  \raction{The answer is: (C)}
\end{tcolorbox}

\section{Additional Results}
\label{additional-results}


\begin{table}[h]
\centering
\scriptsize
\setlength{\tabcolsep}{2pt}
\renewcommand{\arraystretch}{1.1}
\caption{Results on the SeedBench-Avg split.}
\label{tab:seedbench_avg_split}
\resizebox{\linewidth}{!}{%
\begin{tabular}{ll*{10}{S[table-format=3.*]}}
\toprule
\textbf{VLM} & \textbf{Setting} &
\multicolumn{1}{c}{\makecell{\textbf{Overall}}} &
\multicolumn{1}{c}{\makecell{\textbf{Scene}\\\textbf{Understanding}}} &
\multicolumn{1}{c}{\makecell{\textbf{Instance}\\\textbf{Identity}}} &
\multicolumn{1}{c}{\makecell{\textbf{Instance}\\\textbf{Attributes}}} &
\multicolumn{1}{c}{\makecell{\textbf{Instance}\\\textbf{Location}}} &
\multicolumn{1}{c}{\makecell{\textbf{Instance}\\\textbf{Counting}}} &
\multicolumn{1}{c}{\makecell{\textbf{Spatial}\\\textbf{Relation}}} &
\multicolumn{1}{c}{\makecell{\textbf{Instance}\\\textbf{Interaction}}} &
\multicolumn{1}{c}{\makecell{\textbf{Visual}\\\textbf{Reasoning}}} &
\multicolumn{1}{c}{\makecell{\textbf{Text}\\\textbf{Understanding}}} 
\\
\midrule
\multirow{7}{*}{\shortstack[c]{Qwen2.5-\\VL}}
  & E2E (base)          & 75.28 & 81.00 & 75.00 & 76.00 & 81.00 & 82.00 & 59.00 & 76.29 & 71.00 & 76.47 \\
  & E2E (base + CoT)    & 77.55 & 83.00 & 85.00 & 78.00 & 76.00 & 79.00 & 62.00 & 75.26 & 79.00 & 81.18 \\
  & E2E (SFT)               & 75.63 & 81.00 & 79.00 & 76.00 & 73.00 & 80.00 & 58.00 & 79.38 & 77.00 & 77.65 \\
  & E2E (RL)                & 77.89 & 84.00 & 82.00 & 78.00 & 77.00 & 81.00 & 58.00 & 77.32 & 82.00 & 82.35 \\
  & \method{} (base) & 68.93 & 76.00 & 80.00 & 75.00 & 68.00 & 79.00 & 57.00 & 56.70 & 61.00 & 67.06 \\
  & \method{} (SFT)      & 66.55 & 75.00 & 70.00 & 71.00 & 62.00 & 73.00 & 58.00 & 59.79 & 60.00 & 70.59 \\
  & \method{} (RL)       & 76.42 & 79.00 & 87.00 & 78.00 & 72.00 & 81.00 & 70.00 & 70.10 & 74.00 & 76.47 \\
\midrule
\multirow{6}{*}{\shortstack[c]{Llama3.2-\\Vision}}
  & E2E (base)          & 73.24 & 77.00 & 80.00 & 77.00 & 74.00 & 68.00 & 52.00 & 74.23 & 79.00 & 78.82 \\
  & E2E (base + CoT)    & 74.72 & 78.00 & 83.00 & 80.00 & 75.00 & 72.00 & 56.00 & 75.26 & 76.00 & 77.65 \\
  & E2E (SFT)               & 68.03 & 76.00 & 75.00 & 76.00 & 58.00 & 62.00 & 43.00 & 71.13 & 76.00 & 76.47 \\
  & \method{} (base) & 66.33 & 73.00 & 75.00 & 77.00 & 67.00 & 65.00 & 55.00 & 62.89 & 62.00 & 58.82 \\
  & \method{} (SFT)      & 65.31 & 67.00 & 78.00 & 75.00 & 65.00 & 72.00 & 50.00 & 53.61 & 63.00 & 63.53 \\
  & \method{} (RL)       & 75.17 & 85.00 & 81.00 & 80.00 & 74.00 & 77.00 & 60.00 & 69.07 & 72.00 & 78.82 \\
\midrule
\multicolumn{2}{l}{Number of questions} &
\multicolumn{1}{c}{882} &
\multicolumn{1}{c}{100} &
\multicolumn{1}{c}{100} &
\multicolumn{1}{c}{100} &
\multicolumn{1}{c}{100} &
\multicolumn{1}{c}{100} &
\multicolumn{1}{c}{100} &
\multicolumn{1}{c}{97} &
\multicolumn{1}{c}{100} &
\multicolumn{1}{c}{85} \\
\bottomrule
\end{tabular}%
}
\end{table}

\end{document}